\documentclass{article}

\usepackage{microtype}
\usepackage{graphicx}
\usepackage{subfigure}
\usepackage{booktabs} 
\usepackage{amsmath}
\usepackage{amssymb}

\usepackage{hyperref}

\usepackage{enumitem}

\newcommand{\bs}[1]{\boldsymbol{#1}}
\newcommand{\CD}{\mathcal{D}}
\newcommand{\EE}{\mathbb{E}}
\newcommand{\zv}{\boldsymbol{z}}
\newcommand{\thetav}{\boldsymbol{\theta}}
\newcommand{\xv}{\boldsymbol{x}}
\newcommand{\phiv}{\boldsymbol{\phi}}
\newcommand{\epsilonv}{\boldsymbol{\epsilon}}

\usepackage[accepted]{icml2018}

\icmltitlerunning{Adversarial Time-to-Event Modeling}

\begin{document}

\twocolumn[
\icmltitle{Adversarial Time-to-Event Modeling}




\begin{icmlauthorlist}
	\icmlauthor{Paidamoyo Chapfuwa}{to}
	\icmlauthor{Chenyang  Tao}{to}
	\icmlauthor{Chunyuan  Li }{to}
	\icmlauthor{Courtney  Page}{to}
	\icmlauthor{Benjamin  Goldstein}{to}
	\icmlauthor{Lawrence  Carin}{to}
	\icmlauthor{Ricardo  Henao}{to}
	
\end{icmlauthorlist}

\icmlaffiliation{to}{Duke University}
\icmlcorrespondingauthor{Paidamoyo Chapfuwa}{paidamoyo.chapfuwa@duke.edu}

\icmlkeywords{Machine Learning, ICML}

\vskip 0.3in
]



\printAffiliationsAndNotice{}  

\begin{abstract}
	Modern health data science applications leverage abundant molecular and electronic health data, providing opportunities for machine learning to build statistical models to support clinical practice.
	Time-to-event analysis, also called survival analysis, stands as one of the most representative examples of such statistical models.
	We present a deep-network-based approach that leverages adversarial learning to address a key challenge in modern time-to-event modeling: nonparametric estimation of event-time distributions.
	We also introduce a principled cost function to exploit information from censored events (events that occur subsequent to the observation window).
	Unlike most time-to-event models, we focus on the estimation of time-to-event distributions, rather than time ordering.
	We validate our model on both benchmark and real datasets, demonstrating that the proposed formulation yields significant performance gains relative to a parametric alternative, which we also propose.
\end{abstract}

\section{Introduction}
Time-to-event modeling is one of the most widely used statistical analysis tools in biostatistics and, more broadly, health data science applications.
For a given subject, these models estimate either a risk score or the {time-to-event} distribution, from a pre-specified point in time at which a set of covariates (predictors) are observed.
In practice, the model is parameterized as a weighted, often linear, combination of covariates. Time is estimated parametrically or nonparametrically, the former by assuming an underlying time distribution and the latter as proportional to observed event times.
These models have been widely used in risk profiling \cite{hippisley2013predicting,cheng2013development}, treatment planning, and drug development \cite{fischl1987efficacy}.
Time-to-event modeling, and in a larger context, point processes, constitute the fundamental analytical tools in applications for which the future behavior of a system or individual is to be characterized statistically.

The principal time-to-event modeling tool is the Cox Proportional Hazards (Cox-PH) model \cite{cox1992regression}.
Cox-PH is a semi-parametric model that assumes the effect of covariates is a fixed, time-independent, multiplicative factor on the hazard rate, which characterizes the instantaneous {death} rate of the surviving population.
By optimizing a partial likelihood formulation, Cox-PH circumvents the difficulty of specifying the unknown, time-dependent, baseline hazard function.
Consequently, Cox-PH results in point-estimates \emph{proportional} to the event times.
Further, estimation of Cox-PH models depends heavily on event ordering and not the time-to-event itself, which is known to compromise the scalability of the estimation procedure to large datasets.
This poor scaling behavior is manifested because the formulation is not amenable to stochastic training with minibatches.

It is well accepted that the fixed-covariate-effects assumption made in Cox-PH is strong, and unlikely to hold in reality \cite{aalen1994effects}.
For instance, individual heterogeneity and other sources of variation, often likely to be dependent on time, are rarely measured or totally unobservable.
This unobservable variation has been gradually recognized as a major concern in survival analysis and cannot be safely ignored \cite{collett2015modelling,aalen2001understanding}.
When these sources of variation are independent of time, they can be modeled via fixed or random effects \cite{aalen1994effects,hougaard1995frailty}.
However, in cases for which they render the hazard rate time-dependent, such variation is difficult to control, diagnose, or model parametrically.
Cox-PH is known to be sensitive to such assumption violations \cite{aalen2001understanding,kleinbaum2010survival}.
Moreover, Cox-PH focuses on the estimation of the covariate effects rather than the survival time distribution, \emph{i.e.}, time-to-event prediction.
The motivation behind Cox-PH and its shortcomings make it less appealing in applications where prediction is of highest importance.

An alternative to the Cox-PH model is the Accelerated Failure Time (AFT) model \cite{wei1992accelerated}.
AFT makes the simplifying assumption that the effect of covariates either accelerates or delays the event progression, relative to a \emph{parametric} baseline time-to-event distribution.
However, by not making the baseline hazard a constant, as in standard Cox-PH, AFT is often a more reasonable assumption in clinical settings when predictions are important \cite{wei1992accelerated}.
AFT also encompasses a wide range of popular \emph{parametric} proportional hazards models and proportional odds models, when the event baseline time distribution is specified properly \cite{klein2005survival}.
Learning in AFT models falls into the category of maximum likelihood estimation, and
therefore it scales well to large datasets, when trained via stochastic gradient descent.
Further, AFT is also more robust to unobserved variation effects, relative to Cox-PH \cite{keiding1997role}.

From a machine learning perspective, recent advances in deep learning are starting to transform clinical practice.
Equipped with modern learning techniques and abundant data, machine-learning-driven diagnostic applications have surpassed human-expert performance in a wide array of health care applications \citep{cheng2013development,cheng2016computer,havaei2017brain,djuric2017precision,gulshan2016development}.
However, applications involving time-to-event modeling have been largely under-explored.
From the existing approaches, most focus on extending Cox-PH with nonlinear neural-network-based covariate mappings \citep{katzman2016deep,faraggi1995neural,zhu2016deep}, casting the time-to-event modeling as a discretized-time classification problem \citep{yu2011learning, fotso2018deep}, or introducing a nonlinear map between covariates and time via Gaussian processes \citep{fernandez2016gaussian,alaa2017gaussianmulti}. 
Interestingly, all of these approaches focus their applications toward relative risk, fixed-time risk (\emph{e.g.}, 1-year mortality) or competing events, rather than event-time estimation, which is key to \emph{individualized} risk assessment.

Generative Adversarial Networks (GANs) \citep{goodfellow2014generative} have recently demonstrated unprecedented potential for generative modeling, in settings where the goal is to estimate complex data distributions via implicit sampling.
This is done by specifying a flexible generator function, usually a deep neural network, whose samples are adversarially optimized to match in distribution to those from real data.
Succesful examples of GAN include generation of images \citep{radford2015unsupervised,salimans2016improved}, text  \citep{yu2017seqgan,zhang2017adversarial} and data conditioned on covariates \citep{isola2016image,reed2016generative}.
However, ideas from adversarial learning are yet to be exploited for the challenging task that is time-to-event modeling.

Previous work often represents time-to-event distributions using a limited family of parametric forms, \emph{i.e.}, log-normal, Weibull, Gamma, Exponential, \emph{etc}.
It is well understood that parametric assumptions are often violated in practice, largely because of the model is unable to capture unobserved (nuisance) variation.
This fundamental shortcoming is one of the main reasons why non-parametric methods, \emph{e.g.}, Cox proportional hazards, are so popular.
Adversarial learning leverages a representation that implicitly specifies a time-to-event distribution via sampling, rather than learning the parameters of a pre-specified distribution.
Further, GAN-learning penalizes unrealistic samples, which is a known issue in likelihood-based models \cite{karras2018progressive}.

The work presented here seeks to improve the quality of the predictions in nonparametric time-to-event models.
We propose a deep-network-based nonparametric time-to-event model called a \emph{Deep Adversarial Time-to-Event} (DATE) model.
Unlike existing approaches, DATE focuses on the estimation of time-to-event distributions, rather than event ordering, thus emphasizing predictive ability.
Further, this is done while accounting for missing values, high-dimensional data and censored events.
The key contributions associated with the DATE model are: ($i$) The first application of GANs to nonlinear and nonparametric time-to-event modeling, conditioned on covariates.
($ii$) A principled censored-event-aware cost function that is distribution-free and independent of time ordering.
($iii$) Improved uncertainty estimation via deep neural networks with stochastic layers.
($iv$) An alternative, parametric, non-adversarial time-to-event AFT model to be used as baseline in our experiments.
($v$) Results on benchmark and real data demonstrate that DATE outperforms its parametric counterpart by a substantial margin.
%

\section{Background}
%
Time-to-event datasets are usually composed of three variables $\{\xv_i,t_i,l_i\}_{i=1}^N=\CD$, the covariates $\xv_i = [x_{i1}, ..., x_{ip}] \in \mathbb{R}^{p}$ and event pairs $(t_{i}, l_{i})$, where $t_{i}$ is the time-to-event of interest and $l_{i} \in \{0, 1\}$ is a binary censoring indicator.
Typically, $l_{i}=1$ indicates the event is observed, alternatively, $l_{i} = 0$ indicates censoring at $t_i$ (when $l_{i} = 0$, an event, if it occurs, transpires after the observation window that ends at time $t_i$).
$N$ denotes the size of the dataset.

Time-to-event models characterize the survival function:
\begin{align*}
S(t) = \ P(T>t) = \ 1-F(t) = \exp\left(-\textstyle{\int}_{0} ^{t} h(s) ds\right) \,,
\end{align*}
defined as the fraction of the population that survives up to time $t$, or the \emph{conditional} hazards rate function $h(t|\xv)$ (see below), defined as the instantaneous rate of occurrence of an event at time $t$ given covariates $\xv$.
We derive the relationship between $h(t|\xv)$ and $S(t)$ from standard definitions \cite{kleinbaum2010survival}:
\begin{align}\label{eq:hazardfunc}
h(t|\xv) = \lim\limits_{dt \rightarrow 0}  \frac{P(t <T < t + dt|\xv)}{P(T > t|\xv) dt} = \frac{f(t|\xv)}{S(t|\xv)} \,,
\end{align}

where $f(t|\xv)$ is the conditional survival density function and $S(t|\xv)$ is formally the complement of the cumulative conditional density function $F(t|\xv)$.
To solve the hazard-rate differential equation \eqref{eq:hazardfunc}, we establish the following relationship between $f(t|\xv)$, $h(t|\xv)$ and $S(t|\xv)$ via the cumulative conditional hazard $H(t|\xv) = \int_{0}^{t} h(s|\xv) ds$ as
\begin{align}
S(t|\xv) & = \exp\left(-H(t|\xv)\right) \,, \label{eq:survivalfunc} \\
\label{eq:survivaldist}
f(t|\xv) & = h(t|\xv)S(t|\xv) \,.
\end{align}
See Supplementary Material for examples of parametric time-to-event characterizations.
Time-to-event models, \emph{e.g.}, Cox-PH and AFT, leverage the results in \eqref{eq:survivalfunc} and \eqref{eq:survivaldist}, to characterize the relationship between covariates $\xv$ and time-to-event $t$, when estimating the conditional hazard function $h(t|\xv)$.
Two popular frameworks, Cox-PH and AFT, approach the estimation of $h(t| \xv)$ using nonparametric and parametric techniques, respectively.
\paragraph{Cox Proportional Hazard}
The Cox-PH \cite{cox1992regression} model is a semi-parametric, linear model where the conditional hazard function $h(t|\xv)$ depends on time through the baseline hazard $h_{0}(t)$, and independent of covariates $\xv$ as
\begin{align}\label{eq:coxh}
h(t| \xv) 
= & \ h_{0}(t)\exp(\xv^{\top}\bs{\beta}) \,.
\end{align}
Provided with the $N$ observation tuples in ${\cal D}$, Cox-PH estimates the regression coefficients, $\bs{\beta}\in\mathbb{R}^p$, that maximize the partial likelihood \cite{cox1992regression}:
\begin{align}
\label{eq:coxlik}
{\cal L}(\bs{\beta})  &= \prod_{i:l_i=1}^{} \frac{\exp(\xv_{i}^{\top}\bs{\beta})}{\sum_{j: t_{j} \ge t_{i}} \exp(\xv_{j}^{\top}\bs{\beta})} \,,
\end{align}
where ${\cal L}(\bs{\beta})$ is independent of the baseline hazard in \eqref{eq:coxh}.
Note also that \eqref{eq:coxlik} only depends on the ordering of $t_i$, for $i,\ldots,N$, and not their actual values.
Cox-PH is nonparametric in that it estimates the ordering of the events, not their times, thus avoiding the need to specify a distribution for $h_{0}(t)$.
Several techniques have been developed that assume a parametric distribution for $h_{0}(t)$, in order to estimate the actual time-to-event, however, not nearly as widely adopted as standard Cox-PH.
See \citet{bender2005generating} for specifications of $h_{0}(t)$ that result in an exponential, Weibull or Gompertz survival density functions.
For example, when $f(t)=\lambda\exp(-\lambda t)$, \emph{i.e.,} exponential survival density, thus $h_0(t)=\lambda$.

\paragraph{Accelerated Failure Time}
The AFT model \cite{wei1992accelerated} is a popular alternative to the widely used Cox-PH model.
In this model, similar to Cox-PH, it is assumed that $h(t|\xv)) = \psi(\xv) h_0(\psi(\xv) t)$, where $\psi(\xv)$ is the total effect of covariates, $\xv$, usually through a linear relationship $\psi(\xv)=\exp(-\xv^{\top}\bs{\beta})$, where $\bs{\beta}$ represents the regression coefficients.
If the conditional survival density function satisfies $f(t|\xv)=\psi(\xv) f_0(t)$, \emph{i.e.}, $S(t)$ independent of $\xv$ like in Cox-PH, then we can write
%
\begin{align}\label{eq:aft}
\log t = \log(t_0) - \log \psi(\xv) = \xi - \xv^{\top}\bs{\beta} \,,
\end{align}
where $t_0\sim p_0(t)$ is the \emph{unmoderated} time, thus $\xi$ characterizes the baseline survival density distribution.
Note the similarity between \eqref{eq:coxh} and \eqref{eq:aft} despite the differences in their motivation.
%
%
Different choices of baseline distribution yield a variety of
AFT distributions, including Weibull, log-normal, gamma and inverse Gaussian \cite{klein2005survival}.
Intuitively, AFT assumes the effect of the covariates, $\psi(\xv)$, accelerates or delays the life course, which is often meaningful in a clinical or pharmaceutical setting, and sometimes easier to interpret compared with Cox-PH \cite{wei1992accelerated}.
Empirical evidence has shown that AFT is more robust to missing values and misspecification of the survival function than Cox-PH \cite{keiding1997role}.
Gaussian-process-based AFT models \cite{fine1999proportional,alaa2017gaussianmulti} have been used to model competing risk applications with success.

\section{Adversarial Time-to-Event}
%
We develop a nonparametric model for $p(t|\xv)$, where $t$ is the (non-censored) time-to-event from the time at which covariates $\xv$ were observed. More precisely, we learn the ability to sample from $p(t|\xv)$ via approximation $q(t|\xv)$.
Further, we do so without specifying a distribution for the marginal (baseline survival distribution), $p_0(t)$, which in AFT is usually assumed log-normal.
Like in Cox-PH and AFT, we assume that $p_0(t)$ is independent of covariates $\xv$.

For censored events, $l_i=0$, we wish the model to have a high likelihood for $p(t>t_i|\xv_i)$, while for non-censored events, $l_i=1$, we wish that the pairs $\{\xv_i,t_i\}$ be consistent with data generated from $p(t|\xv)p_0(\xv)$, where $p_0(\xv)$ is the (empirical) marginal distribution for covariates, from which we can sample but whose explicit form is unknown.

We consider a \emph{conditional} generative adversarial network (GAN) \cite{goodfellow2014generative}, in which we draw approximate samples from $p(t|\xv)$, for $l_i=1$, as
\begin{equation}\label{eq:tx_obs}
t = G_{\thetav}(\xv,\epsilonv;l=1), \quad \epsilonv\sim p_\epsilon(\epsilonv) \,,
\end{equation}
where $p_\epsilon(\epsilonv)$ is a simple distribution, \emph{e.g.}, isotropic Gaussian or uniform (discussed below).
The generator, $G_{\thetav}(\xv,\epsilonv;l=1)$ is a deterministic function of $\xv$ and $\epsilonv$, specified as a deep neural network with model parameters $\thetav$, that \emph{implicitly} defines $q_\theta(t|\xv,l=1)$ in a nonparametric manner.
We explicitly note that $l=1$, to emphasize that all $t$ drawn from this model are event times (non-censored times).
Ideally the pairs $\{\xv,t\}$ manifested from the model in \eqref{eq:tx_obs} are indistinguishable from the observed data $\{\xv,t,l=1\}\in\mathcal{D}$, \emph{i.e.}, the non-censored samples.

Let $\mathcal{D}_{nc}\subset \mathcal{D}$ and $\mathcal{D}_{c}\subset \mathcal{D}$ be the disjoint subsets of non-censored and censored data, respectively.
Given a discriminator function $D_{\phiv}(\xv,t)$ specified as a deep neural network with model parameters $\phiv$.
The cost function based on the non-censored data has the following form:
\begin{align}
\ell_1(\thetav,\phiv;\mathcal{D}_{nc}) = & \ \mathbb{E}_{(t,\xv)\sim p_{nc}} [D(\xv,t)] \label{eq:minimax} \\
+ & \ \mathbb{E}_{\xv\sim p_{nc},\epsilonv\sim p_\epsilon} [1-D(\xv,G_{\thetav}(\xv,\epsilonv;l=1))] \,, \notag
\end{align}
where $p_{nc}(t,\xv)$ is the \emph{empirical} joint distribution responsible for $\mathcal{D}_{nc}$, and the expectation terms are estimated through samples $\{t,\xv\}\sim p_{nc}(t,\xv)$ and $\epsilonv \sim p_\epsilon(\epsilonv)$ only.
We seek to maximize $\ell_1(\thetav,\phiv;\mathcal{D}_{nc})$ wrt discriminator parameters $\phiv$, while seeking to minimize it wrt generator parameters $\thetav$.
For non-censored data, the formulation in \eqref{eq:minimax} is the standard conditional GAN.

We also leverage the censored data $\mathcal{D}_{c}$ to inform the parameters $\thetav$ of generative model $G_{\thetav}(\xv,\epsilonv;l=0)$.
We therefore consider the additional cost function
\begin{align}\label{eq:cens}
\begin{aligned}
& \ell_2(\thetav;\mathcal{D}_{c})=\\
& \hspace{10mm}\mathbb{E}_{(t,\xv)\sim p_{c},\epsilonv\sim p_\epsilon}[\max(0, t-G_{\thetav}(\xv,\epsilonv;l=0 )] \,,
\end{aligned}
\end{align}
where $\max(0,\cdot)$ encodes that $G_{\thetav}(\xv,\epsilonv;l=0 )$ incurs no cost as long as the sampled time is larger than the censoring point.
Further, $p_{c}(t,\xv)$ is the \emph{empirical} joint distribution responsible for $\mathcal{D}_{c}$, from which samples $\{t,\xv\}$ are drawn to approximate the expectation in \eqref{eq:cens}.
Note that $\max(0,\cdot)$ is one of many choices; smoothed or margin-based alternatives may be considered, but are not addressed here, for simplicity.

For cases in which the proportion of observed events is low, the costs in \eqref{eq:minimax} and \eqref{eq:cens} underrepresent the desire that time-to-events must be as close as possible to the ground truth, $t$.
For this purpose, we also impose a distortion loss $d(\cdot,\cdot)$
\begin{equation}\label{eq:trecon}
\ell_3(\thetav;  \CD_{nc}) = \EE_{(t, \xv) \sim p_{nc}} [d(t, G_{\thetav}(\xv,\epsilonv;l=1 ))] \,,
\end{equation}
that penalizes $G_{\thetav}(\xv,\epsilonv;l=1)$ for not being close to the event time $t$ for non-censored events only.
In the experiments, we set $d(a,b)=\|a-b\|_1$.

The complete cost function is
\begin{align}\label{eq:cost}
\begin{aligned}
\ell(\thetav,\phiv;\mathcal{D}) = & \ \ell_1(\thetav,\phiv;\mathcal{D}_{nc}) \\
+ & \ \lambda_2\ell_2(\thetav;\mathcal{D}_{c}) + \lambda_3\ell_3(\thetav;\CD_{nc}) \,,
\end{aligned}
\end{align}
where $\{\lambda_2,\lambda_3\} > 0$ are tuning parameters controlling the trade-off between non-censored and censored cost functions relative to the discriminator objective in \eqref{eq:minimax}.
In our experiments we set $\lambda_2=\lambda_3=1$, provided that \eqref{eq:cens} and \eqref{eq:trecon} are written in terms of expectations, thus already account for the proportion differences in $\mathcal{D}_{c}$ and $\mathcal{D}_{nc}$.
However, this may not be sufficient in heavily imbalanced cases or when the time domains for $\mathcal{D}_{c}$ and $\mathcal{D}_{nc}$ are very different.

The cost function in \eqref{eq:cost} is optimized using stochastic gradient descent on minibatches from $\mathcal{D}$. We maximize $\ell(\thetav,\phiv;\mathcal{D})$ wrt $\phiv$ and minimize it wrt $\thetav$. The terms $\ell_1(\thetav,\phiv;\mathcal{D}_{nc})$ and $\ell_3(\thetav;\CD_{nc})$ reward $G_{\thetav}(\xv,\epsilonv;l=1)$ if it synthesizes data that are consistent $\mathcal{D}_{nc}$, and the term $\ell_2(\thetav;\mathcal{D}_{c})$ encourages $G_{\thetav}(\xv,\epsilonv;l=1)$ to generate event times that are consistent with the data $\mathcal{D}_{c}$, \emph{i.e.}, larger than censoring times.
%

\begin{figure}[t!]
	\centering
	\includegraphics[width=0.9\columnwidth]{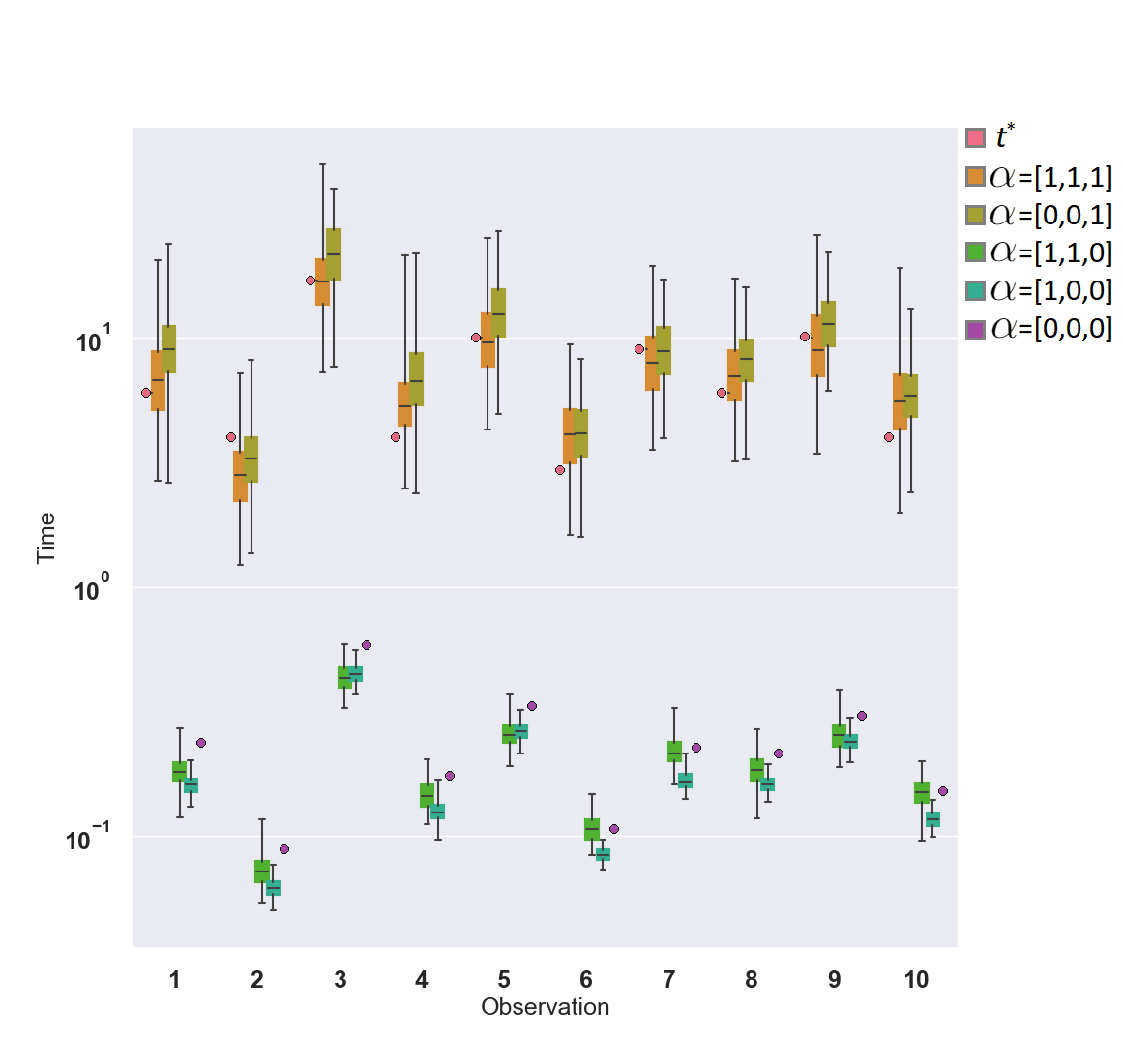}
	\vspace{-6mm}
	\caption{\small Effects of stochastic layers on uncertainty estimation on 10 randomly selected test-set subjects from the {\sc support} dataset. Ground truth times are denoted as $t^*$ and box plots represent time-to-event distributions from a 2-layer model, where $\bs{\alpha}=[\alpha_0,\alpha_1,\alpha_2]$ indicates whether the corresponding noise source, $\{\epsilonv_0,\epsilonv_1, \epsilonv_2\}$, is active. For example $\bs{\alpha}=[1,0,0]$ indicates noise on the input layer only.}
	\label{fg:noise}
	\vspace{-4mm}
\end{figure}

\paragraph{Time-to-event uncertainty}
The generator in \eqref{eq:tx_obs} has a single source of stochasticity, $\epsilonv$, which in GAN-based models has been traditionally applied as input to the model, independent of covariates $\xv$.
In a Multi-Layer Perceptron (MLP) architecture, $\bs{h}_1=g(\bs{W}_{10}\xv+\bs{W}_{11}\epsilonv)$, where $\bs{h}_1$ denotes the vector of layer-1 hidden units, $g(\cdot)$ is the activation function (RELU in the experiments), $\bs{W}_{10}$ and $\bs{W}_{11}$ are weight matrices for covariates and noise, respectively, and we have omitted the bias term for clarity.

In a model with multiple layers, the noise term applied to the input tends to have a small effect on the distribution of sampled event times (see experiments).
More specifically, samples from $q_\theta(t|\xv)$ tend to have small variance.
This results in a model with underestimated uncertainty, hence overconfident predictions.
This is due to many factors, including compounding effects of activation nonlinearities, layer-wise regularizers (\emph{e.g.}, dropout), and cancelling terms when the support of the noise distribution is the real line (both positive and negative).
Although the cost function in \eqref{eq:cost} rewards the generator for producing (non-censored) event times close to the ground-truth, thus in principle encouraging event time distributions to cover it, this rarely happens in practice.
This issue is well-known in the GAN literature \citep{salimans2016improved}.

Here we take a simple approach consisting on adding sources of stochasticity to every layer of the generator as
%
$
\bs{h}_j=g(\bs{W}_{j0}\bs{h}_{j-1}+\bs{W}_{j1}\epsilonv_j) \,,
$
%
where $j=1,\ldots,L$ and $L$ is the number of layers.
By doing this, we encourage increased coverage on the event times produced by the generator, without substantially changing the model or the learning procedure.
In the experiments, we use a multivariate uniform distribution, $\epsilonv_j\sim{\rm Uniform}(0,1)$, for $j=1,\ldots,L$, over Gaussian to reduce cancelling effects.
As we show empirically, this approach produces substantially better coverage compared to having noise only on the input layer and without the convergence issues associated with the additional stochasticity.

In Figure \ref{fg:noise} we illustrate the contribution of the noise on each layer to the distribution of event times.
In this example, we show 10 test-set estimated time-to-event distributions using a 2-layer model with noise sources in all layers, including the input.
We see that ground-truth times are nicely covered by the estimated distributions.
Also, that the combination of noise sources, rather than any individual source, jointly contribute to the desired distribution coverage.
Additional examples, including censored times, are found in the Supplementary Material.

\section{Baseline AFT Model}
%
%
In the experiments, we will demonstrate the capabilities of the DATE model. However, since there are not many scalable nonlinear time-to-event models focused on event time estimation, below we present a parametric (non-adversarial) model to be used in the experiments as baseline.
We do so with the goal of providing a fair comparison and, at the same time, to highlight the distinguishing factors of the DATE model.
Starting from \eqref{eq:aft}, we consider the following MLP-based log-normal AFT:
\begin{align}\label{eq:aftlogn}
\log t = \mu_{\bs{\beta}}(\xv) + \xi \,, \quad \xi \sim {\cal N}(0,\sigma^2_{\bs{\beta}}(\xv)) \,,
\end{align}
where $\mu_{\bs{\beta}}(\xv)$ and $\sigma^2_{\bs{\beta}}(\xv)$ are MLPs parameterized by $\bs{\beta}$, representing the mean and variance of the log-transformed time-to-event as a function of covariates $\xv$.
For convenience, we adopt the log-normal distribution for event time $t$, mainly because we found that it is considerably more stable during optimization, compared to other popular survival distributions, \emph{e.g.}, Weibull or Gamma.

Note that \eqref{eq:aftlogn} is very different from \eqref{eq:tx_obs}, where the generator implicitly defines the distribution $q_\theta(t|\xv)$.
The likelihood function of the log-normal AFT model in \eqref{eq:aftlogn} for all events (censored and non-censored) is then
\vspace{-1mm}
\begin{align}
\prod_i^N p(t_i |\xv_i) = & \prod_{i:l_i = 1} f_{\bs{\beta}}(t_i| \xv_i) \prod_{i:l_i = 0} S_{\bs{\beta}}(t_i |\xv_i) \label{eq:aft_lik} \\
= & \prod_{i:l_i=1} \phi(\nu(t_i,\xv_i)) \prod_{i:l_i=0} (1-\Phi(\nu(t_i,\xv_i))) \,, \nonumber
\end{align}
where $\phi(\cdot)$ and $\Phi(\cdot)$ are the Gaussian density and cumulative density functions, respectively, and $\nu(t_i,\xv_i)=(\log t_i - \mu_{\bs{\beta}}(\xv_i))/\sigma_{\bs{\beta}}(\xv_i)$.
The likelihood in \eqref{eq:aft_lik} is convenient, because it allows estimation of time-to-event, while seemly accounting for censored events.
The latter comes as benefit of having a parametric model with closed-form cumulative density function.

The cost function ${\cal L}( \bs{\beta}; \CD)$ for the Deep Regularized AFT (DRAFT) model is written as:
\begin{equation}\label{eq:loss}
{\cal L}(\bs{\beta};\CD) = \log p(t |{\cal D}) + \eta R(\bs{\beta}; \CD)  \,,
\end{equation}
where the first term is the negative log-likelihood loss from \eqref{eq:aft_lik}, $\eta > 0$ is a tuning parameters, and $R(\bs{\beta}; \CD)$ is a regularization loss that encourages event times to be properly ordered.
Specifically, we use the following cost function adapted from \citet{steck2008ranking}:
\begin{align*}
R(\bs{\beta}; \CD) = \frac{1}{|{\cal E}|}\sum_{i:l_{i}=1} \sum_{j:t_{j} > t_{i}} 1 + \frac{\log \sigma(\mu_{\bs{\beta}}(\xv_j)- \mu_{\bs{\beta}}(\xv_i))}{\log 2} \,,
\end{align*}
where ${\cal E}$ is the set of all pairs $\{i,j\}$ in $\{1,\ldots,N\}$ for which the second argument is observed, \emph{i.e.}, $l_i=1$, and $\sigma(\cdot)$ is the sigmoid function.
The cost function above is a lower bound on the Concordance Index (CI) \cite{harrell1984regression}, which constitutes a difficult-to-optimize discrete objective, that is widely used as performance metric for survival analysis, precisely because it captures time-to-event order.
Further, it is reminiscent of the partial-likelihood of Cox-PH in \eqref{eq:coxlik}, but is more amenable to stochastic training.
\section{Related Work}
%
Deep learning models, specifically MLPs, have been successfully integrated with Cox-PH-based objectives to improve risk estimation in time-to-event models.
\citet{faraggi1995neural} proposed an neural-network-based model optimized using the standard partial-likelihood cost function from Cox-PH.
\citet{katzman2016deep} is similar to \citet{faraggi1995neural}, but leverages modern deep learning techniques such as weight decay, batch normalization and dropout.
\citet{luck2017deep} replaced the partial-likelihood formulation in \eqref{eq:coxlik} with Efron's approximation \citep{efron1977efficiency} and an isotonic regression cost function adapted from \citet{menon2012predicting} to handle censored events.
\citet{zhu2016deep} proposed a time-to-event model for image covariates based on convolutional networks.

From the Gaussian process literature, \citet{fernandez2016gaussian} proposed a time-to-event model inspired by a Poisson process, where the nonlinear map between covariates and time is modeled as a Gaussian process on the Poisson rate.
More recently, \citet{alaa2017gaussianmulti} proposed a deep multi-task Gaussian process model for survival analysis with competing risks, and learned via variational inference.
Following a different path, other approaches recast the time-to-event problem as a classification task.
\citet{yu2011learning} proposed a linear model where (discretized) time is estimated using a sequence of dependent regressors.
More recently, \citet{fotso2018deep} extended their approach to a nonlinear mapping of covariates using deep neural networks.
Generative approaches have also been proposed to infer survival-time distributions with variational inference.
Deep Survival Analysis (DSA) \cite{ranganath2016deep} specifies a latent model that leverages deep exponential family distribution, however their approach does not handle censored events.

All the above methods focus on relative risk quantified as the CI on the ordering of event times, or fixed-time risk, \emph{e.g.}, 1-year mortality.
However, relative risk is most useful when associated with covariate effects, which is difficult in nonlinear models based either on neural networks or Gaussian processes.
Fixed-time risk, although very useful in practice, can be recast as a classification problem rather than a substantially more complex time-to-event model.
Importantly, none of these approaches consider the task of time-to-event estimation, despite the fact that Gaussian process and generative approaches can be repurposed for such task.


\section{Experiments}
%
%
The loss functions in \eqref{eq:cost} and \eqref{eq:loss} for DATE and DRAFT, respectively, are minimized via stochastic gradient descent.
At test time, we draw 200 samples from \eqref{eq:tx_obs} and \eqref{eq:aftlogn} for DATE and DRAFT, respectively, and use medians for quantitative results requiring point estimates, \emph{i.e.}, $\hat{t}={\rm median}(\{t_s\}_{s=1}^{200})$, where $t_s$ is a sample from the trained model.
Detailed network architectures, optimization parameters and initialization settings are in the Supplementary Material.
TensorFlow code to replicate experiments can be found at {\small \url{https://github.com/paidamoyo/adversarial_time_to_event}}.

\vspace{-3mm}
\paragraph{Comparison Methods} For non-deep learning based models, we considered arguably the two most popular approaches to time-to-event modeling, namely, (regularized) Cox-Efron and RSF.
For deep learning models, we considered DRAFT, which generalizes existing neural-network-based methods by using both a parametric log-normal AFT objective and a non-parametric ordering cost function.
Extending DRAFT to a mixture of log-normal distributions with different variances but shared mean, did not result in improved performance.
We did not consider (variational) models for non-censored events only because learning from censored events is one of the main defining characteristics of time-to-event modeling (otherwise, the model essentially becomes non-negative regression).
Further, in most practical situations the proportion of non-censored events is low, \emph{e.g.}, 24\% in the EHR.

\paragraph{Datasets}
%
Our model is evaluated on 4 diverse datasets: $i$) {\sc flchain}: a public dataset
introduced in a study to determine whether non-clonal serum immunoglobin free light chains are predictive of survival time \cite{dispenzieri2012use}.
$ii$) {\sc support}: a public dataset
introduced in a survival time study of seriously-ill hospitalized adults \cite{knaus1995support}.
$iii$) {\sc seer}: a public dataset
provided by the Surveillance, Epidemiology, and End Results Program.
See \cite{ries2007cancer} for details concerning the definition of the 10-year follow-up breast cancer subcohort used in our experiments.
$iv$) {\sc ehr}: a large study from Duke University Health System centered around inpatient visits due to comorbidities in patients with Type-2 diabetes.

%
\begin{table}[t!]
	\vspace{-2mm}
	\centering
	\begin{small}
		\caption{Summary of datasets used in experiments.}
		\label{tb:data}
		\vspace{1mm}
		\begin{tabular}{lrrrr}
			& {\sc ehr} & {\sc flchain} & {\sc support} & {\sc seer} \\
			\hline
			Events (\%) & 23.9 & 27.5 & 68.1 & 51.0 \\
			$N$ & 394,823 & 7,894 & 9,105 & 68,082 \\
			$p$ (cat) & 729 (106) & 26 (21) & 59 (31) & 789 (771) \\
			NaN (\%) & 1.9 & 2.1 & 12.6 & 23.4 \\
			$t_{\rm max}$ & 365 {\tiny days} & 5,215 {\tiny days} & 2,029 {\tiny days} & 120 {\tiny months}
		\end{tabular}
	\end{small}
	\vspace{-6mm}
\end{table}
The datasets are summarized in Table \ref{tb:data}, where $p$ denotes the number of covariates to be analyzed, after one-hot-encoding for categorical (cat) variables.
\emph{Events} indicates the proportion of the observed events, \emph{i.e.}, those for which $l_i = 1$.
NAN indicates the proportion of missing entries in the $N \times p$ covariate matrix and $t_{\rm max}$ is the time range for both censored and non-censored events.

Details about the public datasets: {\sc flchain}, {\sc support} and {\sc seer}, including preprocessing procedures, can be found in the references provided above.
{\sc EHR} is a study designed to track primary care encounters of $19,064$ Type-2 diabetes patients over a period of 10 years (2007-2017).
The purpose of the analysis is to predict diabetes-related causes of hospitalization within 1 year of an EHR-recorded primary care encounter.
Data is processed and analyzed at the patient encounter-level.
The total number of encounters is $N=394,823$.
To avoid bias due to multiple encounters per patient, we split the training, validation and test sets so that a given patient can only be in one of the sets.
The covariates, collected over a period of a year before the primary care encounter of interest, consist of a mixture of continuous and categorical summaries extracted from electronic health records: vitals and labs (minimum, maximum, count and mean values); comorbidities, medications and procedures ICD-9/10 codes (binary indicators and counts); and demographics (age, gender, race, language, smoking indicator, type of insurance coverage, as either continuous or categorical variables).

\begin{figure*}[t!]
	\centering
	\includegraphics[width=0.48\columnwidth]{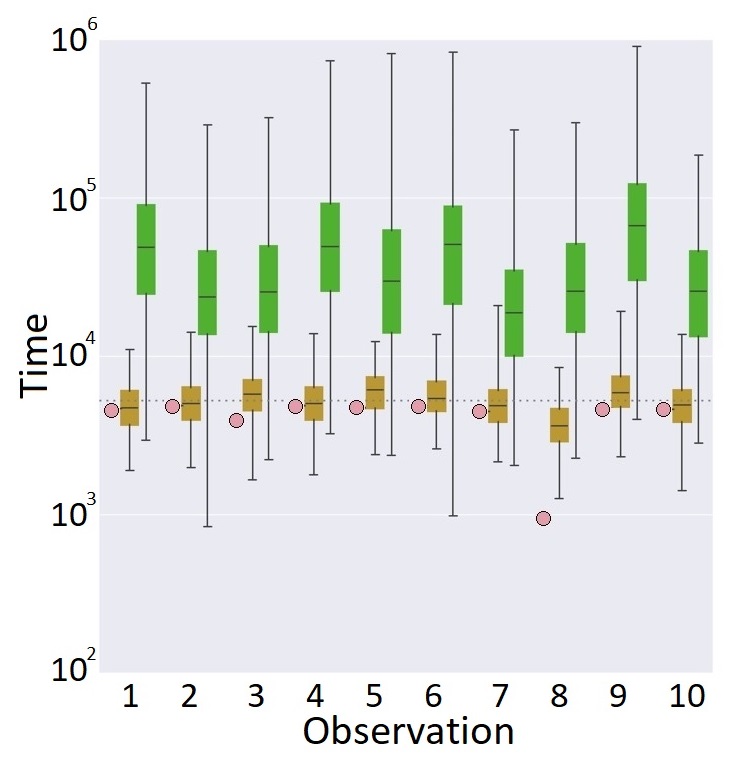}
	\includegraphics[width=0.48\columnwidth]{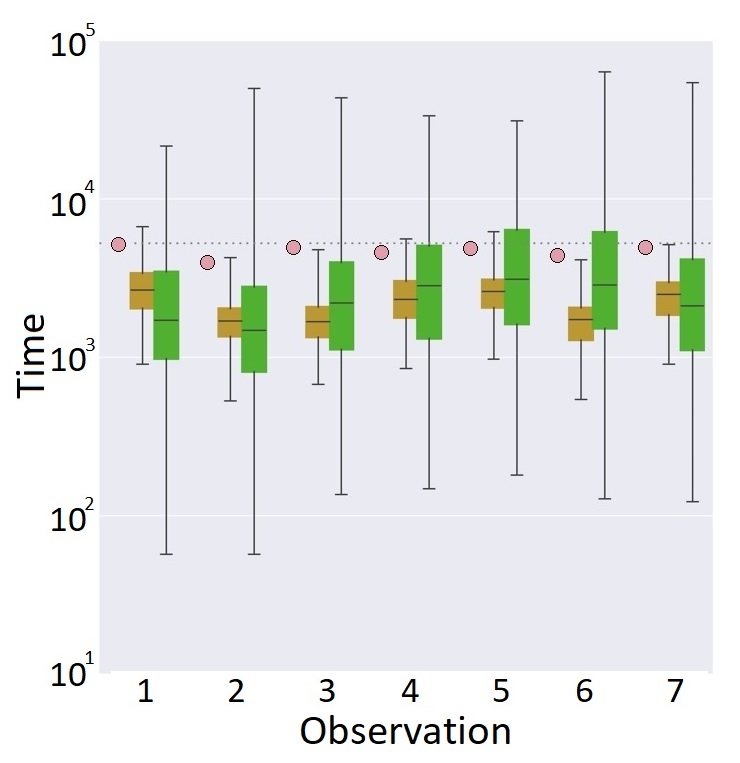}
	\includegraphics[width=0.48\columnwidth]{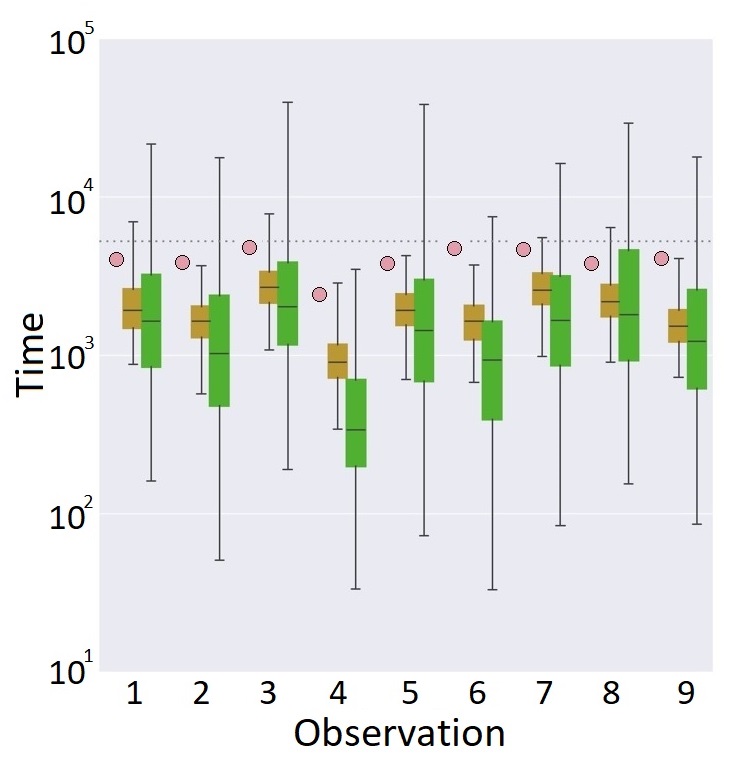}
	\includegraphics[width=0.585\columnwidth]{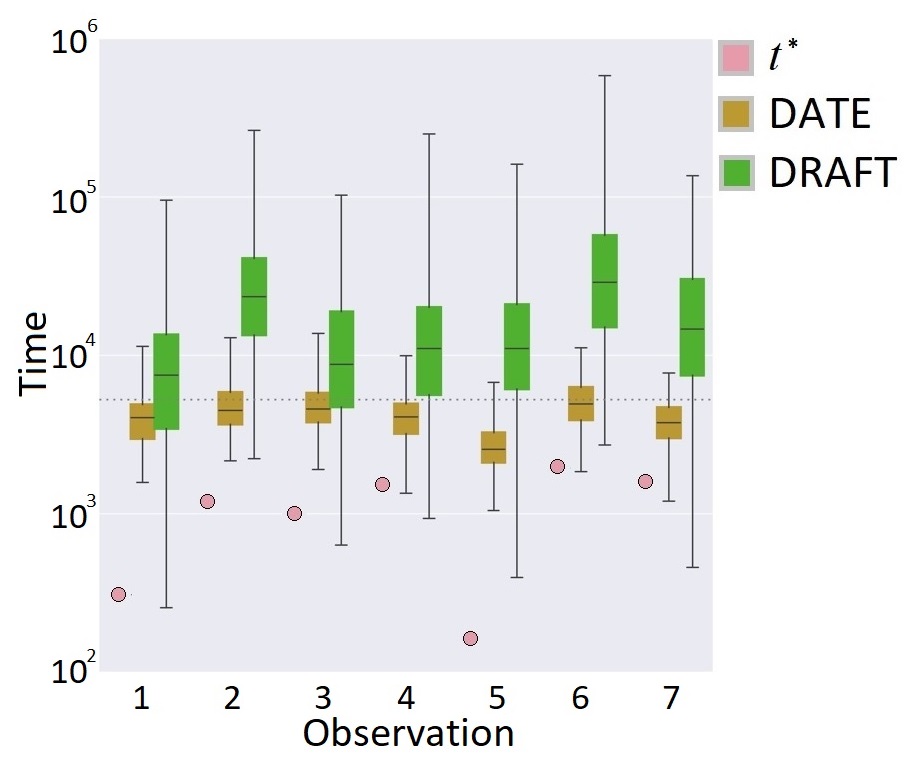}
	\vspace{-4mm}
	\caption{\small Example test-set predictions on {\sc flchain} data. Top best (left) and worst (middle-left) predictions on censored events, and top best (middle-right) and worst (right) predictions on non-censored events.
		Circles denote ground-truth events or censoring points, while box-plots represent distributions over 200 samples for both DATE and DRAFT.
		The horizontal dashed line represents the range ($t_{\rm max}=5,215$ days) of the events.}
	\label{fg:examples}
	\vspace{-4mm}
\end{figure*}

\vspace{-2mm}
\subsection{Qualitative results}
%
First, we visually compare the test-set time-to-event distributions by DATE and DRAFT on {\sc flchain} data.
In Figure \ref{fg:examples} we show the top best (left) and worst (middle-left) predictions on censored events, and the top best (middle-right) and worst (right) predictions on non-censored events.
Circles denote ground-truth events or censoring points, while box-plots represent distributions over 200 samples for both DATE and DRAFT models.
We see that: ($i$) in nearly every case, DATE is more accurate than DRAFT.
($ii$) DRAFT tends to make predictions outside the event range ($t_{\rm max}=5,215$ days), denoted as a horizontal dashed line.
($iii$) DRAFT tends to overestimate the variance of its predictions, approximately by one order of magnitude relative to DATE.
This is not very surprising as DRAFT has an MLP dedicated to estimate, conditioned on the covariates, the variance of the time-to-event distribution.
However, note that variances estimated well over the domain of the events ($t_{\rm max}$) are not necessarily meaningful or desirable.
Figures with similar findings for the other 3 datasets can be found in the Supplementary Material.

\begin{figure}[t!]
	\centering
	\includegraphics[width=0.43\columnwidth,trim=0cm 0mm 2cm 1cm,clip]{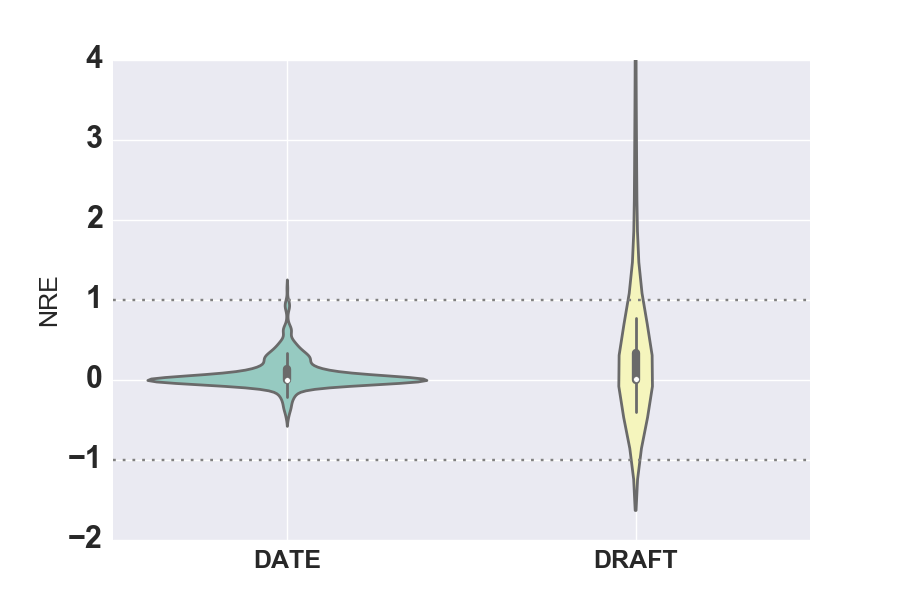}
	\includegraphics[width=0.43\columnwidth,trim=0cm 0mm 2cm 1cm,clip]{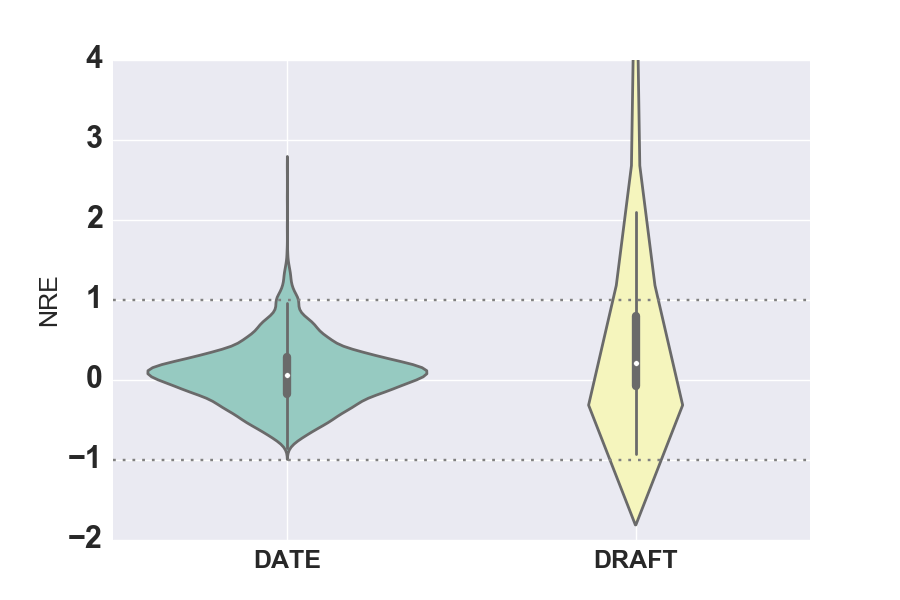}
	\vspace{-2mm}
	\caption{\small Normalized Relative Error (NRE) distribution for {\sc support} (top) and {\sc ehr} (bottom), test-set non-censored events. The horizontal dashed lines represent the range of the events, $t_{\rm max}=120$ months and $t_{\rm max}=365$ days, respectively.}
	\label{fg:boxplots}
	\vspace{-4mm}
\end{figure}

To provide additional insight into the performance of DATE compared to DRAFT, we report the Normalized Relative Error (NRE) defined as $(\hat{t}-t)/t_{\rm max}$ and $\min(0,\hat{t}-t)/t_{\rm max}$ for non-censored and censored events, respectively, where $t$, $\hat{t}$ and $t_{\rm max}$ denote the ground-truth time-to-event, median time estimated (from samples) and event range, as indicated in Table \ref{tb:data}.
The NRE distribution provides a visual representation of the extent of test-set errors, while revealing whether the models are biased toward either overestimating ($\hat{t}>t^*$) or underestimating ($t^*>\hat{t}$) the event times.
Although models with unbiased NREs are naturally preferred, in most clinical applications where being conservative is important, overestimated time-to-events must be avoided as much as possible.
Figure \ref{fg:boxplots} shows NRE distributions for test-set non-censored events on {\sc support} and {\sc ehr} data.
We see that DRAFT results in a considerable amount of errors beyond the event range ($|{\rm NRE}|>1$), $t_{\rm max}=120$ months or $t_{\rm max}=365$ days for {\sc support} and {\sc ehr}, respectively.
Further, we see that the NRE distribution for DRAFT is heavily skewed toward ${\rm NRE}>1$, thus tending to overestimate event times. 
On the other hand, DATE produces errors substantially more concentrated around 0 and within $|{\rm NRE}|<1$, relative to DRAFT.
This demonstrates the advantage of the adversarial method DATE over the likelihood-based method DRAFT in generating realistic samples.
Similar results were observed on the other datasets for both censored and non-censored events.
See Supplementary Material for additional figures.

\vspace{-2mm}
\subsection{Quantitative results}
%
\paragraph{Relative absolute error} The perfromance of DATE is evaluated in terms of absolute error relative to the event range, \emph{i.e.}, $|\hat{t}-t|/t_{\rm max}$.
For censored events, the relative error is defined as $\max(0,t-\hat{t})/t_{\rm max}$, to account for the fact that no error is made as long as $t\leq \hat{t}$.
Table \ref{tb:rae} shows median and 50\% empirical intervals for relative absolute errors on non-censored events, on all test-data.
Results on censored data are small and comparable across approaches, and are thus presented in the Supplementary Material.
Specifically, we see that DATE outperforms DRAFT in 3 our of 4 cases by a substantial margin, and is comparable on the {\sc support} data.
For instance, on {\sc ehr} data 75\% of all DATE test-set predictions have a relative absolute error less than 43\% (approx. 156 days) which is substantially better than the 81\% (approx. 295 days) by DRAFT.

\vspace{-3mm}
\paragraph{Missing data} Since missing data are common in clinical data, \emph{e.g.}, {\sc seer} data contains 23.4\% missing values, we also consider a modified version of DATE, where the generator in \eqref{eq:tx_obs} takes the form $t=G_\theta(\zv,\epsilonv,l=1)$, where $\zv$ is modeled as an adversarial autoencoder \citep{dumoulin2016adversarially,li2017alice, pu2017symmetric} with an encoder/decoder pair specified similar to DATE.
See Supplementary Material for additional details.
This model, denoted in Table \ref{tb:rae} as DATE-AE, does not require missing covariates to be imputed before hand, which is the case of DATE and DRAFT as specified originally.
The results show no substantial performance improvement by DATE-AE, relative to DATE; results indicate that all of these approaches (DRAFT, DATE and DATE-AE) are robust to missing data.
As a benchmark, we took {\sc flchain} and {\sc support}, to then artificially introduced missing values ranging in proportion from 20\% to 50\%.
Results in Supplementary Material support the idea of robustness, since all three approaches resulted in median relative absolute errors within 1\% of those in Table \ref{tb:rae}.

We also tried to quantify statistically the match between time-to-event samples generated from DATE and those from the empirical distribution of the data, using the distribution-free two-sample test based on Maximum Mean Discrepancy (MMD) proposed by \citet{sutherland2016generative}.
Due to sample size limitations (number of non-censored events in the test-set) and high-variances on the $p$-value estimates, we could not reliably reject the hypothesis that real and DATE samples are drawn from the same distribution.
We did confirm it for DRAFT, which is not surprising considering both qualitative and quantitative results discussed above.
\begin{table}[t!]
	\centering
	\caption{Median relative absolute errors (as percentages of $t_{\rm max}$), on non-censored data. Ranges in parentheses are 50\% empirical ranges over (median) test-set predictions.}
	\label{tb:rae}
	\vspace{2mm}
	\resizebox{0.95\columnwidth}{!}{
		\begin{tabular}{lrrr}
			& DATE & DATE-AE & DRAFT \\
			\hline
			{\sc ehr} & {\bf 23.6}$_{(11.1,43.0)}$ & 24.5$_{(12.4,44.0)}$ & 36.7$_{(16.1,81.3)}$ \\
			{\sc flchain} & 19.5$_{(9.5,31.1)}$ & {\bf 19.3}$_{(8.9,32.4)}$ & 26.2$_{(9.0,53.5)}$ \\
			{\sc support} & 2.7$_{(0.4,16.1)}$ & {\bf 1.5}$_{(0.4,19.2)}$ & 2.0$_{(0.2,35.3)}$ \\
			{\sc seer} & {\bf 18.6}$_{(8.3,34.1)}$ & 20.2$_{(10.3,35.8)}$ & 23.7$_{(9.9,51.2)}$
		\end{tabular}
	}
	\vspace{-4mm}
\end{table}

\vspace{-2mm}
\paragraph{Concordance Index} The concordance Index (CI) \cite{harrell1984regression}, which quantifies the degree to which the order of the predicted times is consistent with the ground truth, is the most well-known performance metric in survival analysis.
Although not the focus of our approach, we compared DATE to DATE-AE, DRAFT, Random Survival Forests \cite{ishwaran2008random}, and Cox-PH (with Efron's approximation \cite{efron1977efficiency}).
We found all of these models to be largely comparable.
The results, presented in the Supplementary Material, show DATE(-AE) and DRAFT being the best-performing models on {\sc ehr} and {\sc support}, respectively.
On {\sc seer}, DATE(-AE) and DRAFT outperform Cox-PH and RSF.
Finally, on {\sc flchain}, the smallest dataset, all methods perform about the same.
Note that unlike Cox-PH and DRAFT in \eqref{eq:coxlik} and \eqref{eq:loss}, DATE(-AE) does not explicitly encourage proper time-ordering on the objective function; 
it is consequently deemed a strength of the proposed GAN-based DATE model that it properly learns ordering, without needing to explicitly
impose this condition when training.
DATE does not have a clear advantage in terms of learning the correct order, however, we verified empirically that adding an ordering cost function, $R(\bs{\beta}; \CD)$ in~\eqref{eq:loss}, to DATE does not improve the results.
\begin{table}[t!]
	\centering
	\caption{Median of 95\% intervals for all test-set time-to-event distributions on {\sc support} data. Ranges in parentheses are 50\% empirical quantiles.}
	\label{tb:noise}
	\vspace{2mm}
	\resizebox{\columnwidth}{!}{
		\begin{tabular}{lrrr}
			& Uniform(-1,1) & Uniform(0,1) & Gaussian(0,1) \\
			\hline
			Non-censored & & & \\
			\hline
			All & 60.0$_{(3.9,176.5)}$ & 149.9$_{(8.5,926.8)}$ & 37.9$_{(3.5,237.4)}$ \\
			Input & 28.9$_{(1.8,114.8)}$ & 22.4$_{(1.5,91.2)}$ & 33.7$_{(1.6,127.6)}$ \\
			Output & ---- & 168.8$_{(16.6,844.3)}$ & ---- \\
			\hline
			Censored & & & \\
			\hline
			All & 231.3$_{(177.2,332.1)}$ & 1397.3$_{(990.9,2000.1)}$ & 350.5$_{(254.4,539.3)}$ \\
			Input & 137.3$_{(99.4,205.0)}$ & 86.9$_{(64.4,135.0)}$ & 155.8$_{(106.7,229.3)}$ \\
			Output & ---- & 1158.6$_{(873.8,1670.4)}$ & ---- \\
		\end{tabular}
	}
	\vspace{-5mm}
\end{table}

\vspace{-2mm}
\paragraph{Distribution coverage} We now demonstrate that the DATE model, with noise sources on all layers, has time-to-event distributions with larger variances than versions of DATE with noise only on the input of the neural network.
Table \ref{tb:noise} shows the median of the 95\% intervals for all test-set time-to-event distributions on {\sc support} data.
DATE with Uniform(0,1) has larger variance and coverage compared to the other alternatives, while keeping relative absolute errors and CIs largely unchanged (see Supplementary Material for details).
We did not run models with Uniform(-1,1) and Gaussian(0,1) only on the output layer, because from the other results presented above it is clear that these two options are not nearly as good as having Uniform(0,1) noise on all layers.
Note also that we did not include DRAFT in these comparisons.
DRAFT has naturally good coverage due to the variance of the time-to-event distributions being modeled independent for each observation as a function of the covariates (see for instance Figure \ref{fg:examples}).
However, DRAFT has difficulties keeping good coverage while maintaining good performance, \emph{i.e,}, small absolute relative error. 

\vspace{-2mm}
\section{Conclusions}
%
We have presented an adversarially-learned time-to-event model that leverages a distribution-free cost function for censored events. The proposed approach extends GAN models to time-to-event modeling with censored data, and it is based on deep neural networks with stochastic layers. The model yields improved uncertainty estimation relative to alternative approaches.
As a baseline model for our experiments, we also proposed a parametric AFT-based with a parametric log-normal distribution on the time of event.
To the best of our knowledge, this work is the first to leverage adversarial learning to improve estimation of time-to-event distributions, conditioned on covariates.
Experimental results on challenging time-to-event datasets showed that DATE, our adversarial solution, consistently outperforms DRAFT, its parametric (log-normal) counterpart.
As future work, we will extend DATE to models with competing risks and longitudinally measured covariates.

 \section*{Acknowledgements}
The authors would like to thank the anonymous reviewers for their insightful comments. This research was supported in part by DARPA, DOE, NIH, ONR, NSF and SANOFI. The authors would also like to thank Shuyang Dai, Liqun Chen, Rachel Ballantyne Draelos, Xilin Cecilia Shi, Yan Zhao and Dr. Neha Pagidipati, for fruitful discussions.

\bibliography{date}

\begin{thebibliography}{47}
\providecommand{\natexlab}[1]{#1}
\providecommand{\url}[1]{\texttt{#1}}
\expandafter\ifx\csname urlstyle\endcsname\relax
  \providecommand{\doi}[1]{doi: #1}\else
  \providecommand{\doi}{doi: \begingroup \urlstyle{rm}\Url}\fi

\bibitem[Aalen(1994)]{aalen1994effects}
Aalen, O.~O.
\newblock Effects of frailty in survival analysis.
\newblock \emph{Statistical Methods in Medical Research}, 1994.

\bibitem[Aalen et~al.(2001)Aalen, Gjessing, et~al.]{aalen2001understanding}
Aalen, O.~O., Gjessing, H.~K., et~al.
\newblock Understanding the shape of the hazard rate: A process point of view
  (with comments and a rejoinder by the authors).
\newblock \emph{Statistical Science}, 2001.

\bibitem[Alaa \& van~der Schaar(2017)Alaa and van~der
  Schaar]{alaa2017gaussianmulti}
Alaa, A.~M. and van~der Schaar, M.
\newblock Deep {Multi}-task {Gaussian} {Processes} for {Survival} {Analysis}
  with {Competing} {Risks}.
\newblock In \emph{NIPS}, 2017.

\bibitem[Bender et~al.(2005)Bender, Augustin, and
  Blettner]{bender2005generating}
Bender, R., Augustin, T., and Blettner, M.
\newblock Generating survival times to simulate {Cox} proportional hazards
  models.
\newblock \emph{Statistics in medicine}, 2005.

\bibitem[Cheng et~al.(2016)Cheng, Ni, Chou, Qin, Tiu, Chang, Huang, Shen, and
  Chen]{cheng2016computer}
Cheng, J.-Z., Ni, D., Chou, Y.-H., Qin, J., Tiu, C.-M., Chang, Y.-C., Huang,
  C.-S., Shen, D., and Chen, C.-M.
\newblock Computer-aided diagnosis with deep learning architecture:
  applications to breast lesions in {US} images and pulmonary nodules in {CT}
  scans.
\newblock \emph{Scientific reports}, 2016.

\bibitem[Cheng et~al.(2013)Cheng, Yang, and Anastassiou]{cheng2013development}
Cheng, W.-Y., Yang, T.-H.~O., and Anastassiou, D.
\newblock Development of a prognostic model for breast cancer survival in an
  open challenge environment.
\newblock \emph{Science translational medicine}, 2013.

\bibitem[Collett(2015)]{collett2015modelling}
Collett, D.
\newblock \emph{Modelling survival data in medical research}.
\newblock CRC press, 2015.

\bibitem[Cox(1992)]{cox1992regression}
Cox, D.~R.
\newblock Regression models and life-tables.
\newblock In \emph{Breakthroughs in statistics}. 1992.

\bibitem[Dispenzieri et~al.(2012)Dispenzieri, Katzmann, Kyle, Larson, Therneau,
  Colby, Clark, Mead, Kumar, Melton, et~al.]{dispenzieri2012use}
Dispenzieri, A., Katzmann, J.~A., Kyle, R.~A., Larson, D.~R., Therneau, T.~M.,
  Colby, C.~L., Clark, R.~J., Mead, G.~P., Kumar, S., Melton, L.~J., et~al.
\newblock Use of nonclonal serum immunoglobulin free light chains to predict
  overall survival in the general population.
\newblock In \emph{Mayo Clinic Proceedings}, 2012.

\bibitem[Djuric et~al.(2017)Djuric, Zadeh, Aldape, and
  Diamandis]{djuric2017precision}
Djuric, U., Zadeh, G., Aldape, K., and Diamandis, P.
\newblock Precision histology: how deep learning is poised to revitalize
  histomorphology for personalized cancer care.
\newblock \emph{npj Precision Oncology}, 2017.

\bibitem[Dumoulin et~al.(2016)Dumoulin, Belghazi, Poole, Lamb, Arjovsky,
  Mastropietro, and Courville]{dumoulin2016adversarially}
Dumoulin, V., Belghazi, I., Poole, B., Lamb, A., Arjovsky, M., Mastropietro,
  O., and Courville, A.
\newblock Adversarially learned inference.
\newblock \emph{arXiv}, 2016.

\bibitem[Efron(1977)]{efron1977efficiency}
Efron, B.
\newblock The efficiency of {Cox}'s likelihood function for censored data.
\newblock \emph{JASA}, 1977.

\bibitem[Faraggi \& Simon(1995)Faraggi and Simon]{faraggi1995neural}
Faraggi, D. and Simon, R.
\newblock A neural network model for survival data.
\newblock \emph{Statistics in medicine}, 1995.

\bibitem[Fern{\'a}ndez et~al.(2016)Fern{\'a}ndez, Rivera, and
  Teh]{fernandez2016gaussian}
Fern{\'a}ndez, T., Rivera, N., and Teh, Y.~W.
\newblock Gaussian processes for survival analysis.
\newblock In \emph{NIPS}, 2016.

\bibitem[Fine \& Gray(1999)Fine and Gray]{fine1999proportional}
Fine, J.~P. and Gray, R.~J.
\newblock A proportional hazards model for the subdistribution of a competing
  risk.
\newblock \emph{JASA}, 1999.

\bibitem[Fischl et~al.(1987)Fischl, Richman, Grieco, Gottlieb, Volberding,
  Laskin, Leedom, Groopman, Mildvan, Schooley, et~al.]{fischl1987efficacy}
Fischl, M.~A., Richman, D.~D., Grieco, M.~H., Gottlieb, M.~S., Volberding,
  P.~A., Laskin, O.~L., Leedom, J.~M., Groopman, J.~E., Mildvan, D., Schooley,
  R.~T., et~al.
\newblock The efficacy of azidothymidine ({AZT}) in the treatment of patients
  with {AIDS} and {AIDS}-related complex.
\newblock \emph{New England Journal of Medicine}, 1987.

\bibitem[Fotso(2018)]{fotso2018deep}
Fotso, S.
\newblock Deep neural networks for survival analysis based on a multi-task
  framework.
\newblock \emph{arXiv}, 2018.

\bibitem[Goodfellow et~al.(2014)Goodfellow, Pouget-Abadie, Mirza, Xu,
  Warde-Farley, Ozair, Courville, and Bengio]{goodfellow2014generative}
Goodfellow, I., Pouget-Abadie, J., Mirza, M., Xu, B., Warde-Farley, D., Ozair,
  S., Courville, A., and Bengio, Y.
\newblock Generative adversarial nets.
\newblock In \emph{NIPS}, 2014.

\bibitem[Gulshan et~al.(2016)Gulshan, Peng, Coram, Stumpe, Wu, Narayanaswamy,
  Venugopalan, Widner, Madams, Cuadros, et~al.]{gulshan2016development}
Gulshan, V., Peng, L., Coram, M., Stumpe, M.~C., Wu, D., Narayanaswamy, A.,
  Venugopalan, S., Widner, K., Madams, T., Cuadros, J., et~al.
\newblock Development and validation of a deep learning algorithm for detection
  of diabetic retinopathy in retinal fundus photographs.
\newblock \emph{JAMA}, 2016.

\bibitem[Harrell et~al.(1984)Harrell, Lee, Califf, Pryor, and
  Rosati]{harrell1984regression}
Harrell, F.~E., Lee, K.~L., Califf, R.~M., Pryor, D.~B., and Rosati, R.~A.
\newblock Regression modelling strategies for improved prognostic prediction.
\newblock \emph{Statistics in medicine}, 1984.

\bibitem[Havaei et~al.(2017)Havaei, Davy, Warde-Farley, Biard, Courville,
  Bengio, Pal, Jodoin, and Larochelle]{havaei2017brain}
Havaei, M., Davy, A., Warde-Farley, D., Biard, A., Courville, A., Bengio, Y.,
  Pal, C., Jodoin, P.-M., and Larochelle, H.
\newblock Brain tumor segmentation with deep neural networks.
\newblock \emph{Medical image analysis}, 2017.

\bibitem[Hippisley-Cox \& Coupland(2013)Hippisley-Cox and
  Coupland]{hippisley2013predicting}
Hippisley-Cox, J. and Coupland, C.
\newblock Predicting risk of emergency admission to hospital using primary care
  data: derivation and validation of {QAdmissions} score.
\newblock \emph{BMJ open}, 2013.

\bibitem[Hougaard(1995)]{hougaard1995frailty}
Hougaard, P.
\newblock Frailty models for survival data.
\newblock \emph{Lifetime data analysis}, 1995.

\bibitem[Ishwaran et~al.(2008)Ishwaran, Kogalur, Blackstone, and
  Lauer]{ishwaran2008random}
Ishwaran, H., Kogalur, U.~B., Blackstone, E.~H., and Lauer, M.~S.
\newblock Random survival forests.
\newblock \emph{The annals of applied statistics}, 2008.

\bibitem[Isola et~al.(2016)Isola, Zhu, Zhou, and Efros]{isola2016image}
Isola, P., Zhu, J.-Y., Zhou, T., and Efros, A.~A.
\newblock Image-to-image translation with conditional adversarial networks.
\newblock \emph{arXiv}, 2016.

\bibitem[Karras et~al.(2018)Karras, Aila, Laine, and
  Lehtinen]{karras2018progressive}
Karras, T., Aila, T., Laine, S., and Lehtinen, J.
\newblock Progressive growing of {GANs} for improved quality, stability, and
  variation.
\newblock In \emph{ICLR}, 2018.

\bibitem[Katzman et~al.(2016)Katzman, Shaham, Bates, Cloninger, Jiang, and
  Kluger]{katzman2016deep}
Katzman, J., Shaham, U., Bates, J., Cloninger, A., Jiang, T., and Kluger, Y.
\newblock Deep survival: A deep cox proportional hazards network.
\newblock \emph{arXiv}, 2016.

\bibitem[Keiding et~al.(1997)Keiding, Andersen, and Klein]{keiding1997role}
Keiding, N., Andersen, P.~K., and Klein, J.~P.
\newblock The role of frailty models and accelerated failure time models in
  describing heterogeneity due to omitted covariates.
\newblock \emph{Statistics in medicine}, 1997.

\bibitem[Klein \& Moeschberger(2005)Klein and Moeschberger]{klein2005survival}
Klein, J.~P. and Moeschberger, M.~L.
\newblock \emph{Survival analysis: techniques for censored and truncated data}.
\newblock Springer Science \& Business Media, 2005.

\bibitem[Kleinbaum \& Klein(2010)Kleinbaum and Klein]{kleinbaum2010survival}
Kleinbaum, D.~G. and Klein, M.
\newblock \emph{Survival analysis}.
\newblock Springer, 2010.

\bibitem[Knaus et~al.(1995)Knaus, Harrell, Lynn, Goldman, Phillips, Connors,
  Dawson, Fulkerson, Califf, Desbiens, et~al.]{knaus1995support}
Knaus, W.~A., Harrell, F.~E., Lynn, J., Goldman, L., Phillips, R.~S., Connors,
  A.~F., Dawson, N.~V., Fulkerson, W.~J., Califf, R.~M., Desbiens, N., et~al.
\newblock The {SUPPORT} prognostic model: objective estimates of survival for
  seriously ill hospitalized adults.
\newblock \emph{Annals of internal medicine}, 1995.

\bibitem[Li et~al.(2017)Li, Liu, Chen, Pu, Chen, Henao, and Carin]{li2017alice}
Li, C., Liu, H., Chen, C., Pu, Y., Chen, L., Henao, R., and Carin, L.
\newblock Alice: Towards understanding adversarial learning for joint
  distribution matching.
\newblock In \emph{NIPS}, 2017.

\bibitem[Luck et~al.(2017)Luck, Sylvain, Cardinal, Lodi, and
  Bengio]{luck2017deep}
Luck, M., Sylvain, T., Cardinal, H., Lodi, A., and Bengio, Y.
\newblock Deep {Learning} for {Patient}-{Specific} {Kidney} {Graft} {Survival}
  {Analysis}.
\newblock \emph{arXiv}, 2017.

\bibitem[Menon et~al.(2012)Menon, Jiang, Vembu, Elkan, and
  Ohno-Machado]{menon2012predicting}
Menon, A.~K., Jiang, X.~J., Vembu, S., Elkan, C., and Ohno-Machado, L.
\newblock Predicting accurate probabilities with a ranking loss.
\newblock In \emph{ICML}, 2012.

\bibitem[Pu et~al.(2017)Pu, Chen, Dai, Wang, Li, and Carin]{pu2017symmetric}
Pu, Y., Chen, L., Dai, S., Wang, W., Li, C., and Carin, L.
\newblock Symmetric {V}ariational {A}utoencoder and {C}onnections to
  {A}dversarial {L}earning.
\newblock \emph{arXiv}, 2017.

\bibitem[Radford et~al.(2015)Radford, Metz, and
  Chintala]{radford2015unsupervised}
Radford, A., Metz, L., and Chintala, S.
\newblock Unsupervised representation learning with deep convolutional
  generative adversarial networks.
\newblock \emph{arXiv}, 2015.

\bibitem[Ranganath et~al.(2016)Ranganath, Perotte, Elhadad, and
  Blei]{ranganath2016deep}
Ranganath, R., Perotte, A., Elhadad, N., and Blei, D.
\newblock Deep survival analysis.
\newblock In \emph{Machine Learning for Healthcare Conference}, 2016.

\bibitem[Reed et~al.(2016)Reed, Akata, Yan, Logeswaran, Schiele, and
  Lee]{reed2016generative}
Reed, S., Akata, Z., Yan, X., Logeswaran, L., Schiele, B., and Lee, H.
\newblock Generative adversarial text to image synthesis.
\newblock \emph{arXiv}, 2016.

\bibitem[Ries et~al.(2007)Ries, Young~Jr, Keel, Eisner, Lin, and
  Horner]{ries2007cancer}
Ries, L. A.~G., Young~Jr, J.~L., Keel, G.~E., Eisner, M.~P., Lin, Y.~D., and
  Horner, M.-J.~D.
\newblock Cancer survival among adults: {US SEER} program, 1988--2001.
\newblock \emph{Patient and tumor characteristics SEER Survival Monograph
  Publication}, 2007.

\bibitem[Salimans et~al.(2016)Salimans, Goodfellow, Zaremba, Cheung, Radford,
  and Chen]{salimans2016improved}
Salimans, T., Goodfellow, I., Zaremba, W., Cheung, V., Radford, A., and Chen,
  X.
\newblock Improved techniques for training gans.
\newblock In \emph{NIPS}, 2016.

\bibitem[Steck et~al.(2008)Steck, Krishnapuram, Dehing-oberije, Lambin, and
  Raykar]{steck2008ranking}
Steck, H., Krishnapuram, B., Dehing-oberije, C., Lambin, P., and Raykar, V.~C.
\newblock On ranking in survival analysis: Bounds on the concordance index.
\newblock In \emph{NIPS}, 2008.

\bibitem[Sutherland et~al.(2017)Sutherland, Tung, Strathmann, De, Ramdas,
  Smola, and Gretton]{sutherland2016generative}
Sutherland, D.~J., Tung, H.-Y., Strathmann, H., De, S., Ramdas, A., Smola, A.,
  and Gretton, A.
\newblock Generative models and model criticism via optimized maximum mean
  discrepancy.
\newblock \emph{ICLR}, 2017.

\bibitem[Wei(1992)]{wei1992accelerated}
Wei, L.-J.
\newblock The accelerated failure time model: a useful alternative to the {Cox}
  regression model in survival analysis.
\newblock \emph{Statistics in medicine}, 1992.

\bibitem[Yu et~al.(2011)Yu, Greiner, Lin, and Baracos]{yu2011learning}
Yu, C.-N., Greiner, R., Lin, H.-C., and Baracos, V.
\newblock Learning patient-specific cancer survival distributions as a sequence
  of dependent regressors.
\newblock In \emph{NIPS}, 2011.

\bibitem[Yu et~al.(2017)Yu, Zhang, Wang, and Yu]{yu2017seqgan}
Yu, L., Zhang, W., Wang, J., and Yu, Y.
\newblock Seqgan: Sequence generative adversarial nets with policy gradient.
\newblock In \emph{AAAI}, 2017.

\bibitem[Zhang et~al.(2017)Zhang, Gan, Fan, Chen, Henao, Shen, and
  Carin]{zhang2017adversarial}
Zhang, Y., Gan, Z., Fan, K., Chen, Z., Henao, R., Shen, D., and Carin, L.
\newblock Adversarial feature matching for text generation.
\newblock In \emph{ICML}, 2017.

\bibitem[Zhu et~al.(2016)Zhu, Yao, and Huang]{zhu2016deep}
Zhu, X., Yao, J., and Huang, J.
\newblock Deep convolutional neural network for survival analysis with
  pathological images.
\newblock In \emph{Bioinformatics and Biomedicine (BIBM), 2016 IEEE
  International Conference on}, 2016.

\end{thebibliography}


\begin{thebibliography}{3}
\providecommand{\natexlab}[1]{#1}
\providecommand{\url}[1]{\texttt{#1}}
\expandafter\ifx\csname urlstyle\endcsname\relax
  \providecommand{\doi}[1]{doi: #1}\else
  \providecommand{\doi}{doi: \begingroup \urlstyle{rm}\Url}\fi

\bibitem[Glorot \& Bengio(2010)Glorot and Bengio]{glorot2010understanding}
Glorot, Xavier and Bengio, Yoshua.
\newblock Understanding the difficulty of training deep feedforward neural
  networks.
\newblock In \emph{AISTATS}, 2010.

\bibitem[Ioffe \& Szegedy(2015)Ioffe and Szegedy]{ioffe2015batch}
Ioffe, Sergey and Szegedy, Christian.
\newblock Batch normalization: Accelerating deep network training by reducing
  internal covariate shift.
\newblock In \emph{ICML}, 2015.

\bibitem[Kinga \& Adam(2015)Kinga and Adam]{kinga2015method}
Kinga, D and Adam, J~Ba.
\newblock A method for stochastic optimization.
\newblock In \emph{ICLR}, 2015.

\end{thebibliography}
\bibliographystyle{icml2018}

\end{document}


\appendix

\twocolumn[
\icmltitle{ Supplemental Material for \\``Adversarial Time-to-Event Modeling"}



\vskip 0.3in
]

\section{Missing data and DATE-AE }
DATE-AE extends DATE by jointly learning the mapping $\bs{x}\to\bs{z}\to t$ , where $\bs{z}$ is modeled as an adversarial autoencoder. For imputation, the covariates (entries of $\bs{x}$) in the encoder are set to zero if the entry is missing.
When evaluating the reconstruction loss $\gamma_3$ in \eqref{eq: dasa_ae}, we only do so for observed covariates; in this way the autoencoder can learn the correlation structure of the observed data despite missingness and without the need for imputation, while letting the decoder, $\bs{x}={\rm decoder}(\bs{z})$, handle the imputation if needed.
Note that for time-to-event prediction, at test time, we do not have to impute missing values as we can directly evaluate $\bs{x}\to\bs{z}\to t$. DATE-AE, extends DATE formulation with additional  autoencoder discriminator and generator losses shown below:
	\begin{align}\label{eq: dasa_ae}
\gamma_1(\bs{\theta_x},  \bs{\theta_z}, \bs{\psi}; \CD) &= \EE_{(\bs{x}, \tilde{\bs{z}})} [ D_{\bs{\psi}}(\bs{x}, \tilde{\bs{z}}) ]\notag\\
&+ \EE_{(  \tilde{\bs{x}}, \bs{z}   )} [1 - D_{\bs{\psi}}( \tilde{\bs{x}}, \bs{z} )],   \notag\\
\gamma_2(\bs{\theta_x},  \bs{\theta_z}; \CD) &= \EE_{\left(\bs{z} \sim p(\bs{z}), \hat{\bs{z}}\right)}[d(\bs{z}, \hat{\bs{z}})],  \notag\\
\gamma_3(\bs{\theta_x},  \bs{\theta_z}; \CD) &= \EE_{\left(\bs{x} \sim p(\bs{x}), \hat{\bs{x}}\right)} [d(\bs{x} ,\hat{\bs{x}})], \notag\\
\min_{\bs{\theta_x}, \bs{\theta_z}} \max_{\bs{\psi}}\gamma(\bs{\theta_x}, \bs{\theta_z}, \bs{\psi}; \CD) &= 	\gamma_1(\bs{\theta_x},  \bs{\theta_z}, \bs{\psi}; \CD)  \notag\\
&+ \zeta_2 \gamma_2(\bs{\theta_x},  \bs{\theta_z};\CD) \notag\\
&+	\zeta_3 \gamma_3(\bs{\theta_x},  \bs{\theta_z}; \CD) \,,
\end{align}
where $\bs{x} \sim p(\bs{x})$,  $\bs{\tilde{z}} =G_{\bs{\theta_x}}(\bs{x}, \bs{\epsilon_x})$ , $\bs{z} \sim p(\bs{z})$,  $\bs{\tilde{x}} = G_{\bs{\theta_z}}(\bs{z}, \bs{\epsilon_z})$,  $\bs{\epsilon} $ is the noise source, $d$ is the distortion measure and $\{\zeta_2,\zeta_3\}$ are reconstruction tuning parameters.

Tables \ref{tb:missing_rae} and \ref{tb:missing_ci} compares the effects of randomly introducing missing values on the Flchain relative absolute error and concordance-index  respectively.

\section{Concordance index and relative absolute error}
Tables \ref{tb:c-index} and \ref{tb:rae} show comparisons on concordance-index and relative absolute error across all datasets.

\section{Normalized Relative Error (NRE)}
\label{app: relative_error}
\begin{figure}[h!]\centering
	\vspace{00mm}
	\begin{tabular}{c c}            
		\hspace{-4mm}
		\includegraphics[width=4.08cm]{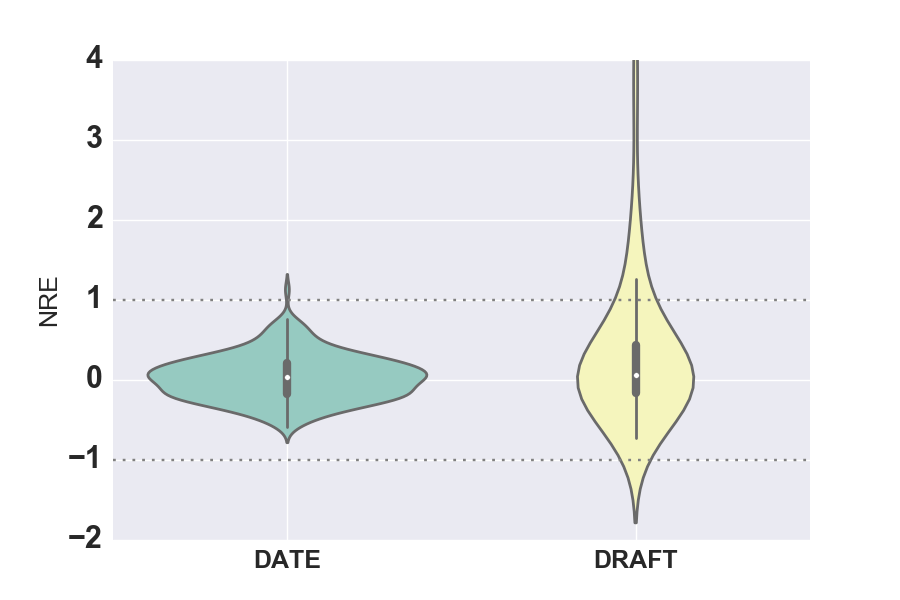}  
		& 
		\hspace{-3mm}
		\includegraphics[width=4.08cm]{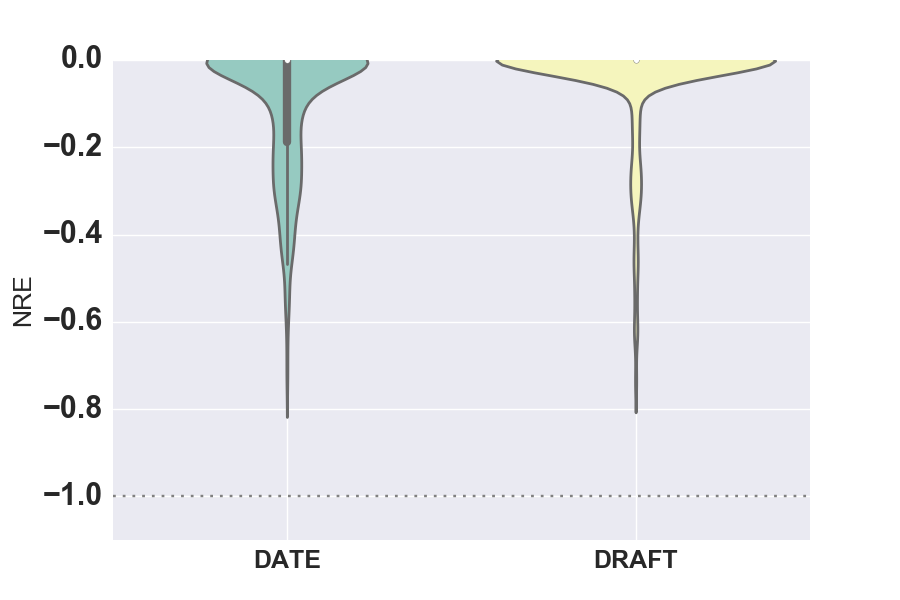}  \\
		(a) Non-Censored \hspace{-0mm} & \hspace{-5mm}
		(b) Censored 
	\end{tabular}
	\vspace{-5mm}
	\caption{\small Normalized relative error on 	{\sc flchain} test data.}
	\vspace{-1mm}
	\label{fig:nre_flchain}
\end{figure}

\begin{figure}[h!]\centering
	\vspace{-2mm}
	\begin{tabular}{c c}            
	\hspace{-4mm}
	\includegraphics[width=4.08cm]{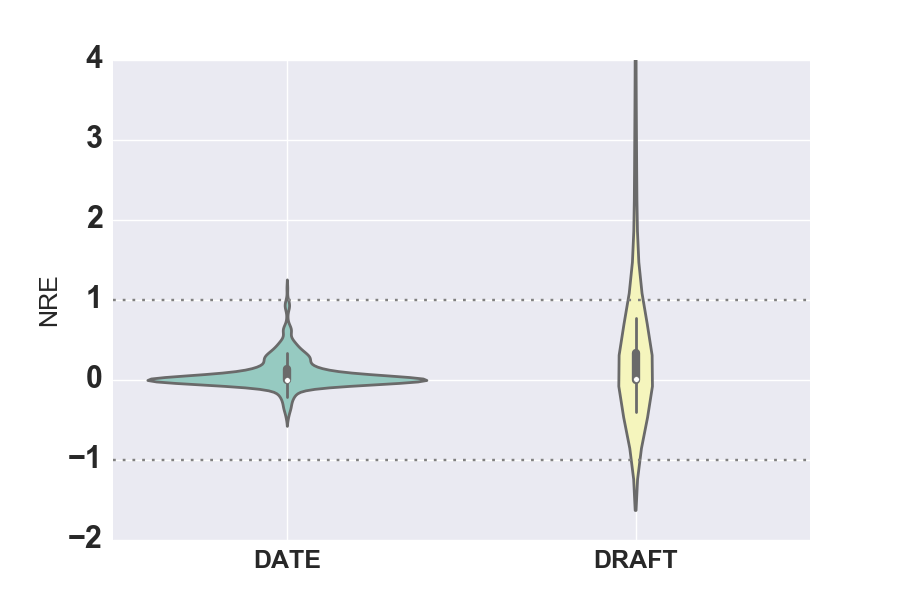}  
	& 
	\hspace{-5mm}
	\includegraphics[width=4.08cm]{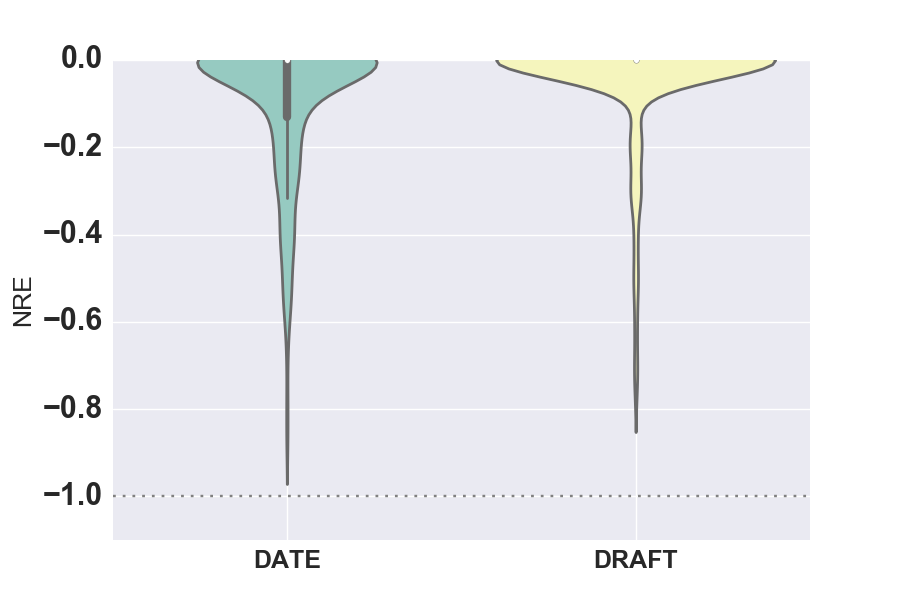}  \\
	(a) Non-Censored \hspace{-0mm} & \hspace{-5mm}
	(b) Censored 
\end{tabular}
\vspace{-5mm}
\caption{\small Normalized relative error on 	{\sc support} test data.}
\vspace{-1mm}
\label{fig:nre_support}
\end{figure}

\begin{figure}[h!] \centering
	\vspace{-0mm}
	\begin{tabular}{c c}            
		\hspace{-4mm}
		\includegraphics[width=4.08cm]{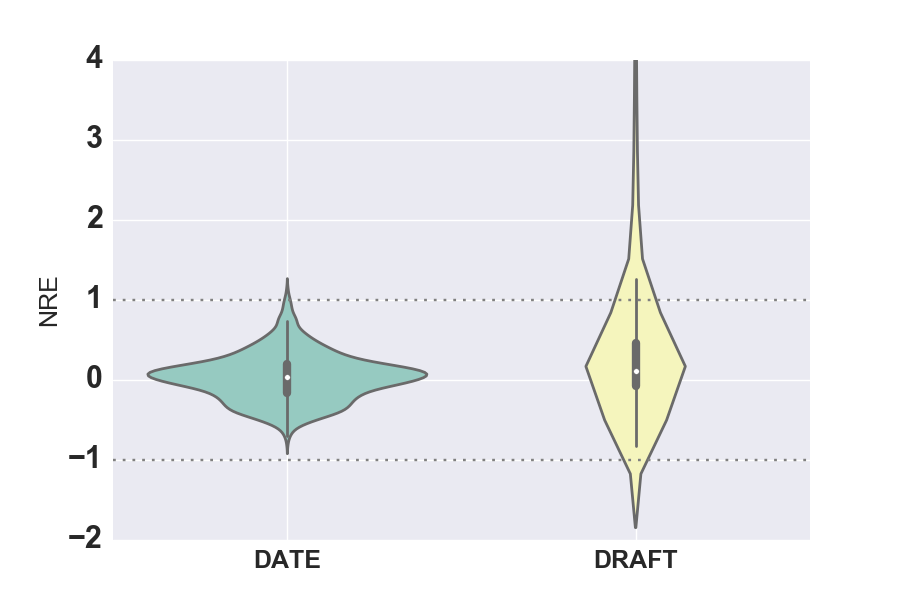}  
		& 
		\hspace{-3mm}
		\includegraphics[width=4.08cm]{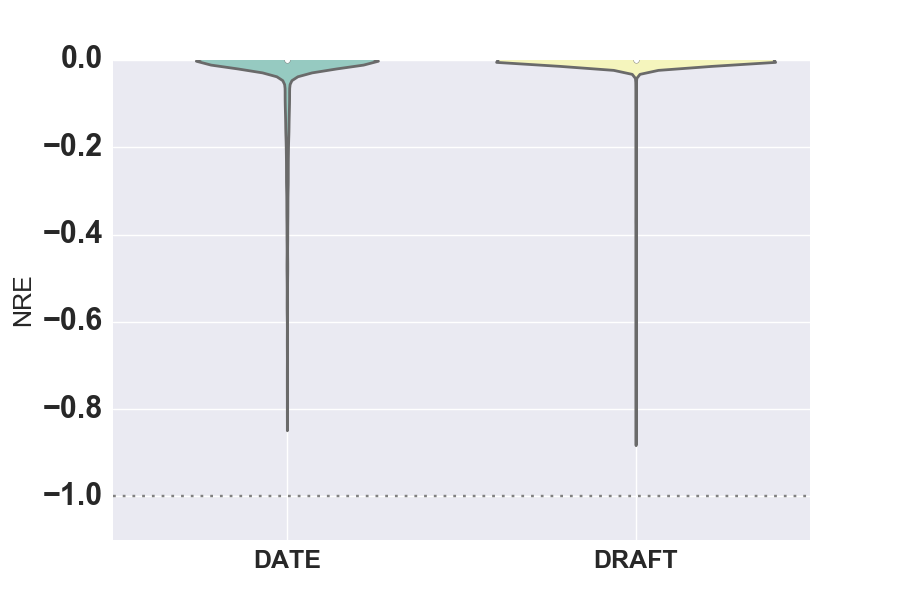}  \\
		(a) Non-Censored \hspace{-0mm} & \hspace{-5mm}
		(b) Censored 
	\end{tabular}
	\vspace{-5mm}
	\caption{\small Normalized relative error on 	{\sc seer} test data.}
	\vspace{-1mm}
	\label{fig:nre_seer}
\end{figure}

\begin{figure}[h!] \centering
	\vspace{-2mm}
	\begin{tabular}{c c}            
		\hspace{-2mm}
		\includegraphics[width=4.08cm]{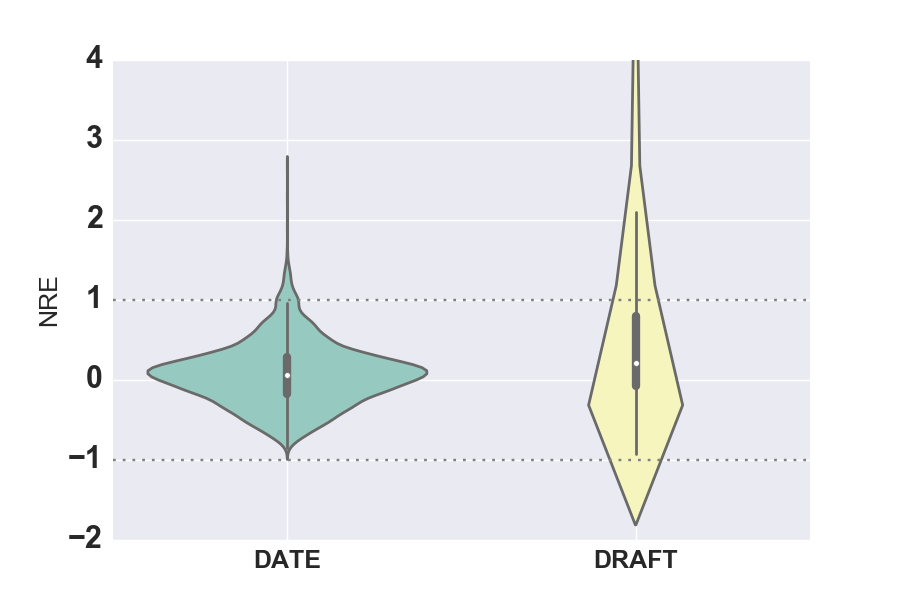}  
		& 
		\hspace{-3mm}
		\includegraphics[width=4.08cm]{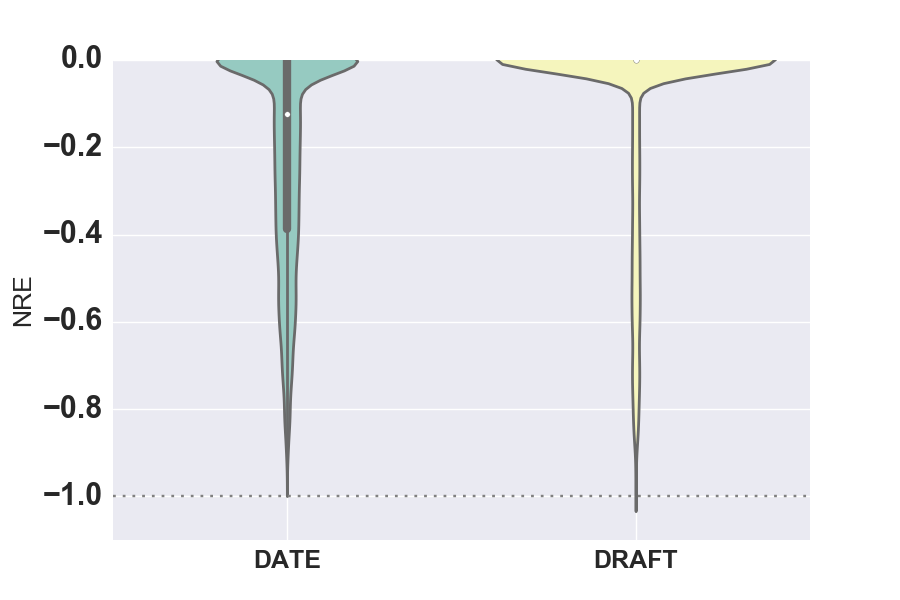}  \\
		(a) Non-Censored \hspace{-0mm} & \hspace{-5mm}
		(b) Censored 
	\end{tabular}
	\vspace{-5mm}
	\caption{\small Normalized relative error on 	{\sc ehr} test data.}
	\vspace{5mm}
	\label{fig:nre_sanofi}
\end{figure}

Figures \ref{fig:nre_flchain},  \ref{fig:nre_support}, \ref{fig:nre_seer} and \ref{fig:nre_sanofi}, show comparison on NRE distributions for both censored and non-censored events.
%
\section{Test set time-to-event distributions}\label{app: samples}
\begin{figure} [h!]\centering
	\vspace{-2mm}
	\includegraphics[width=8cm]{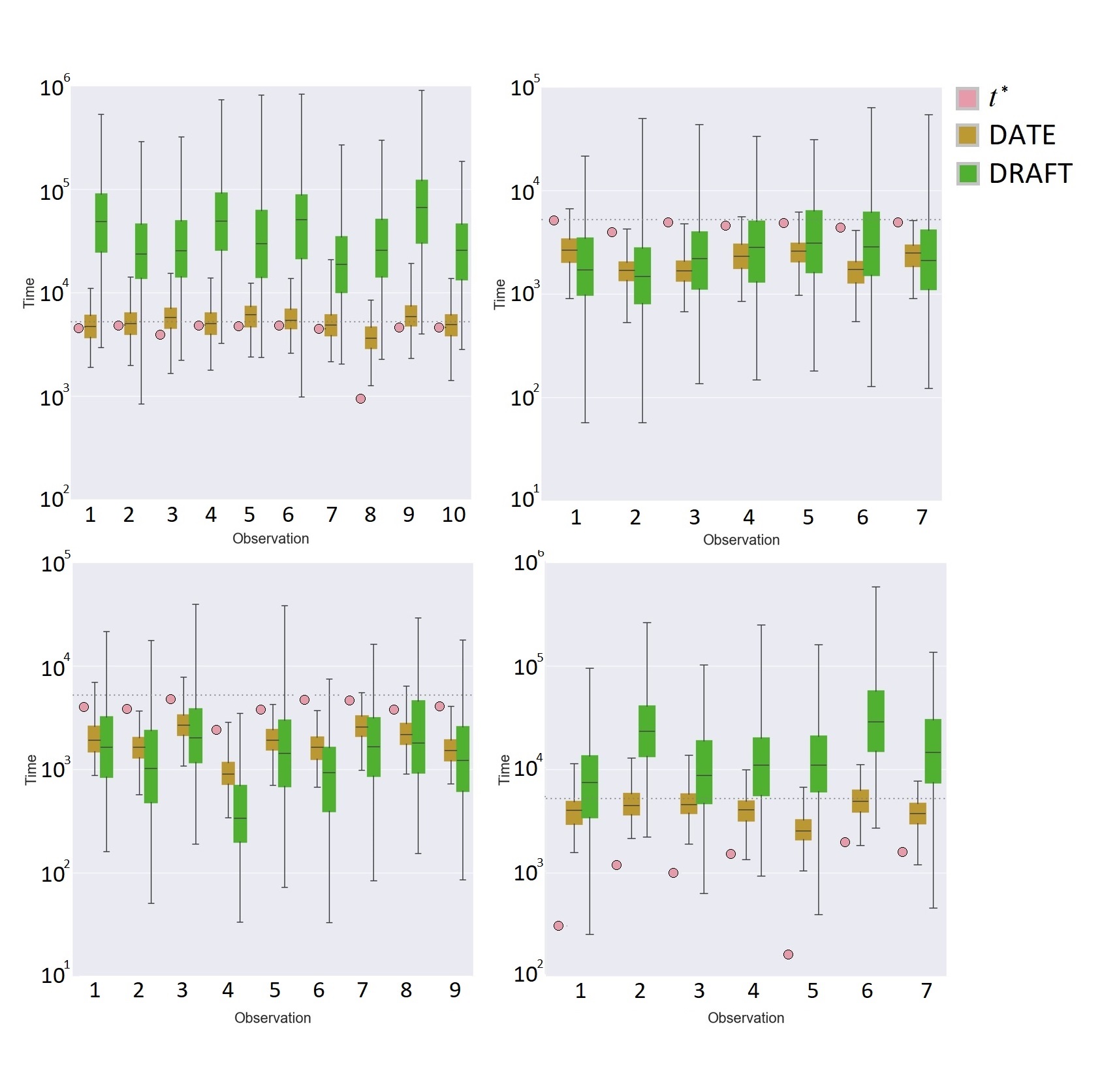}  
	\vspace{-8mm}
	\caption{\small Comparison on {\sc flchain}  Censored best (top-left), worst (top-right) and Non-Censored  best (bottom-left), worst (bottom-right).}
	\label{fig:sam_flchain}
\end{figure}

\begin{figure} [h!]\centering
	\vspace{-4mm}
	\includegraphics[width=8cm]{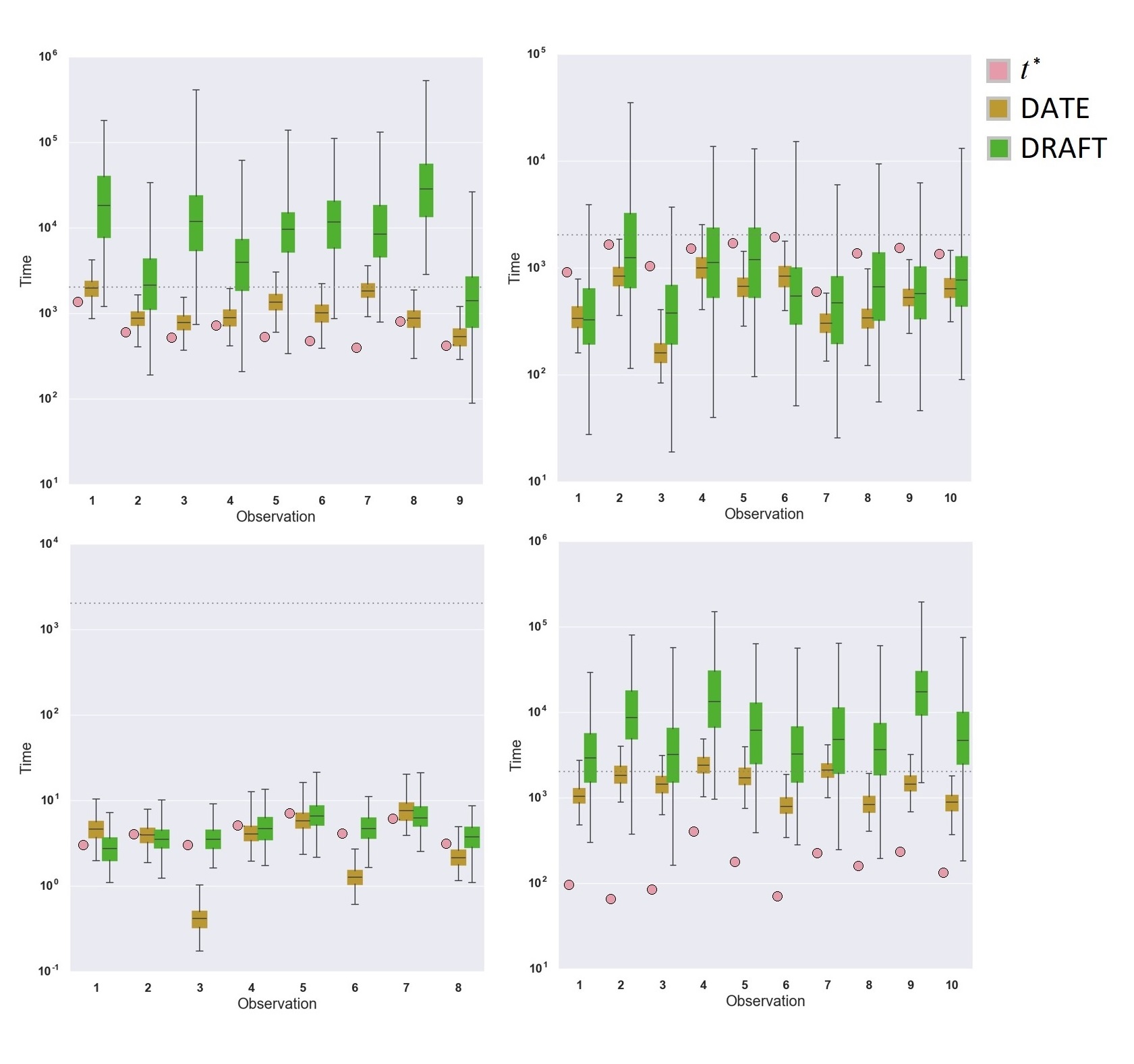}  
		\vspace{-6mm}
		\caption{\small Comparison on {\sc support}  Censored best (top-left), worst (top-right) and Non-Censored  best (bottom-left), worst (bottom-right).}
	\label{fig:sam_support}
\end{figure}

\begin{figure} [h!]\centering
	\vspace{-0mm}
	\includegraphics[width=8cm]{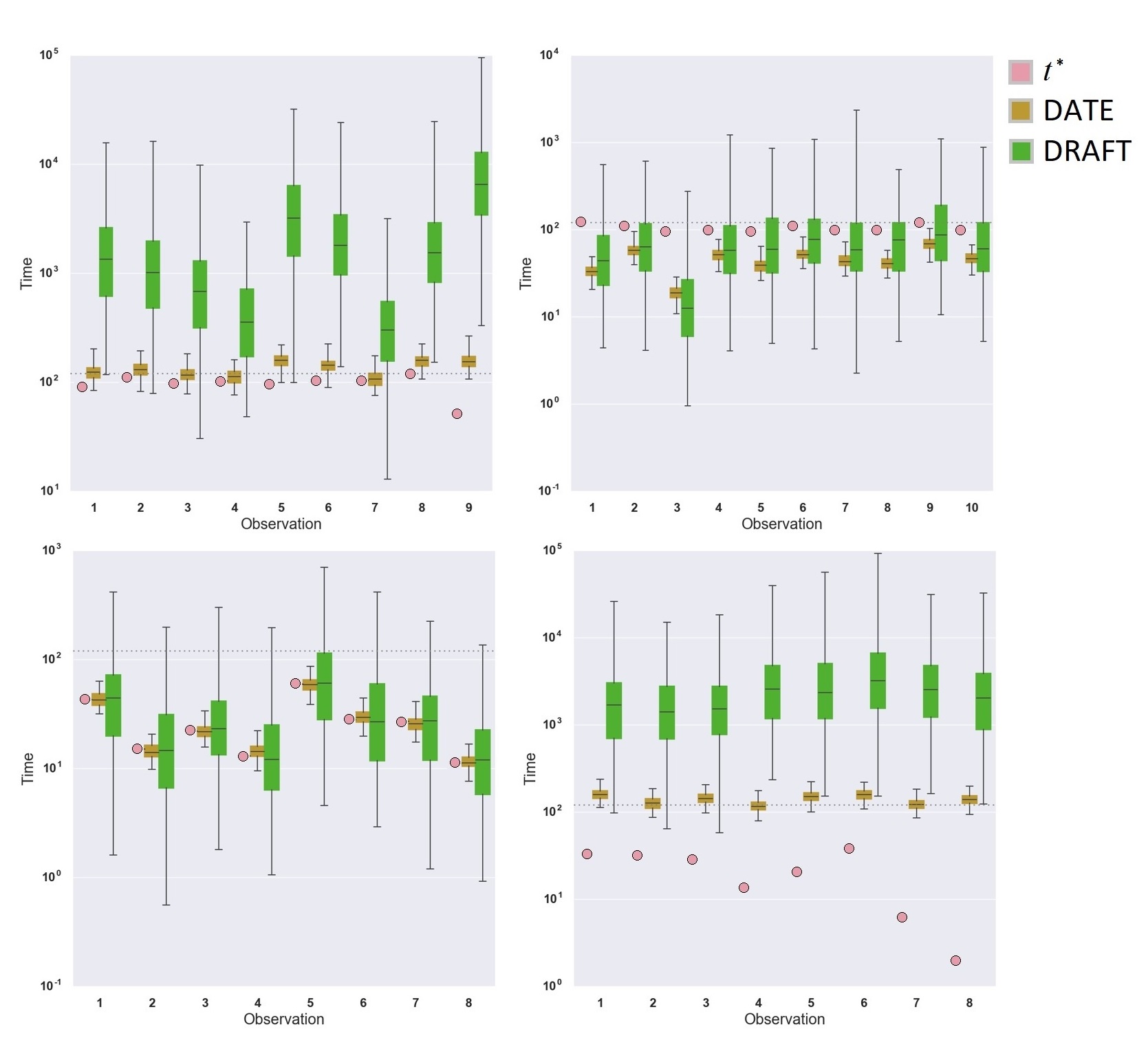}  
		\vspace{-4mm}
		\caption{\small Comparison on {\sc seer}  Censored best (top-left), worst (top-right) and Non-Censored  best (bottom-left), worst (bottom-right).}
	\label{fig:sam_seer}
\end{figure}

\begin{figure} [h!]\centering
	\vspace{-4mm}
	\includegraphics[width=8cm]{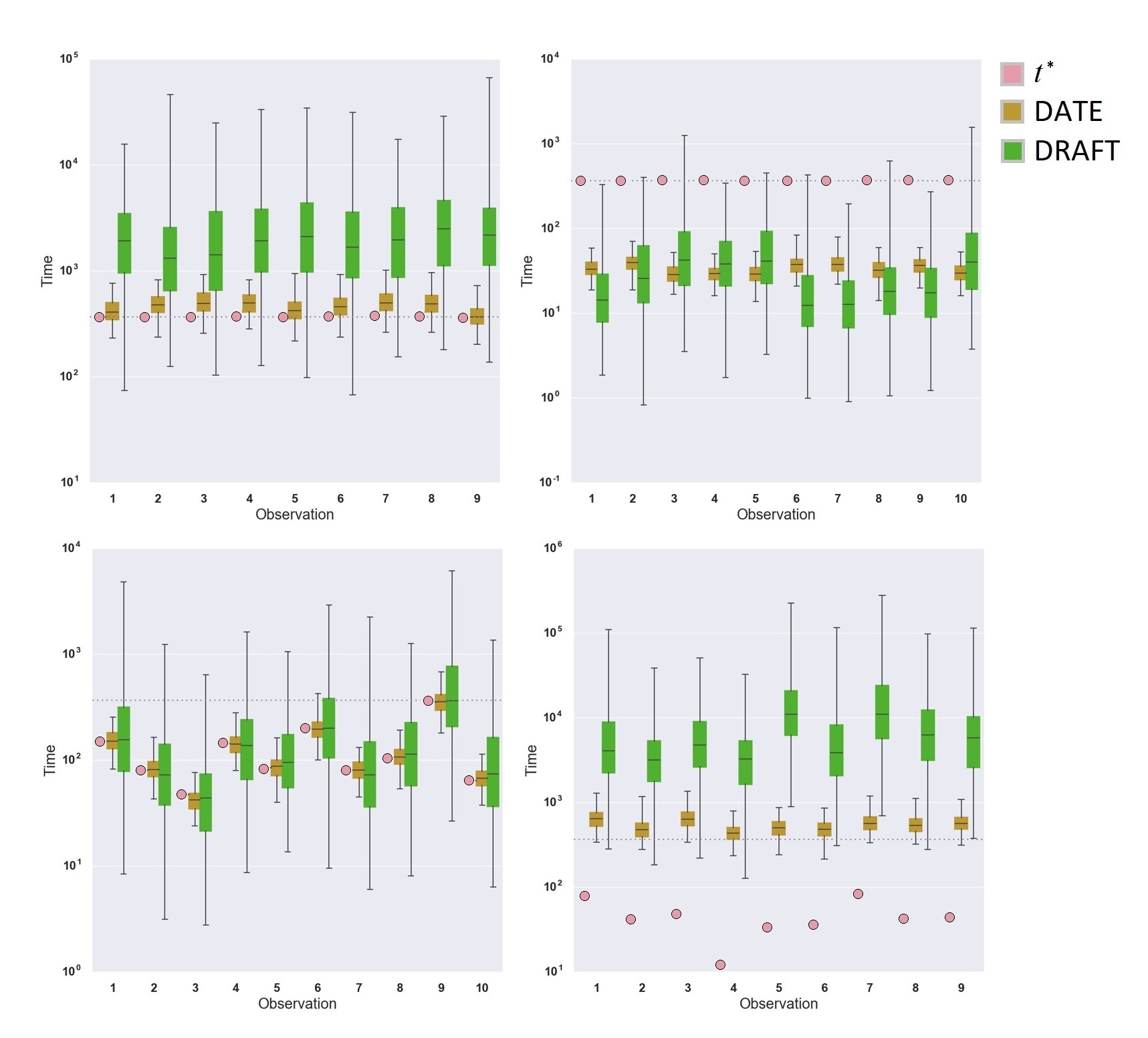}  
		\vspace{-4mm}
	\caption{\small Comparison on {\sc ehr}  Censored best (top-left), worst (top-right) and Non-Censored  best (bottom-left), worst (bottom-right).}
	\label{fig:sam_sanofi}
\end{figure}

We randomly draw best and worst observation samples based on the NRE metric. Figures \ref{fig:sam_flchain} , \ref{fig:sam_support}, \ref{fig:sam_seer} and \ref{fig:sam_sanofi}, show the corresponding distributions comparisons relative to the ground truth or censored time $t^{\star}$.

\section{Effects of  noise source and stochastic layers}
\begin{figure*}[t!]  \centering
	\vspace{-0mm}
	\begin{tabular}{c c c c}            
		\hspace{-4mm}
		\includegraphics[width=4.08cm]{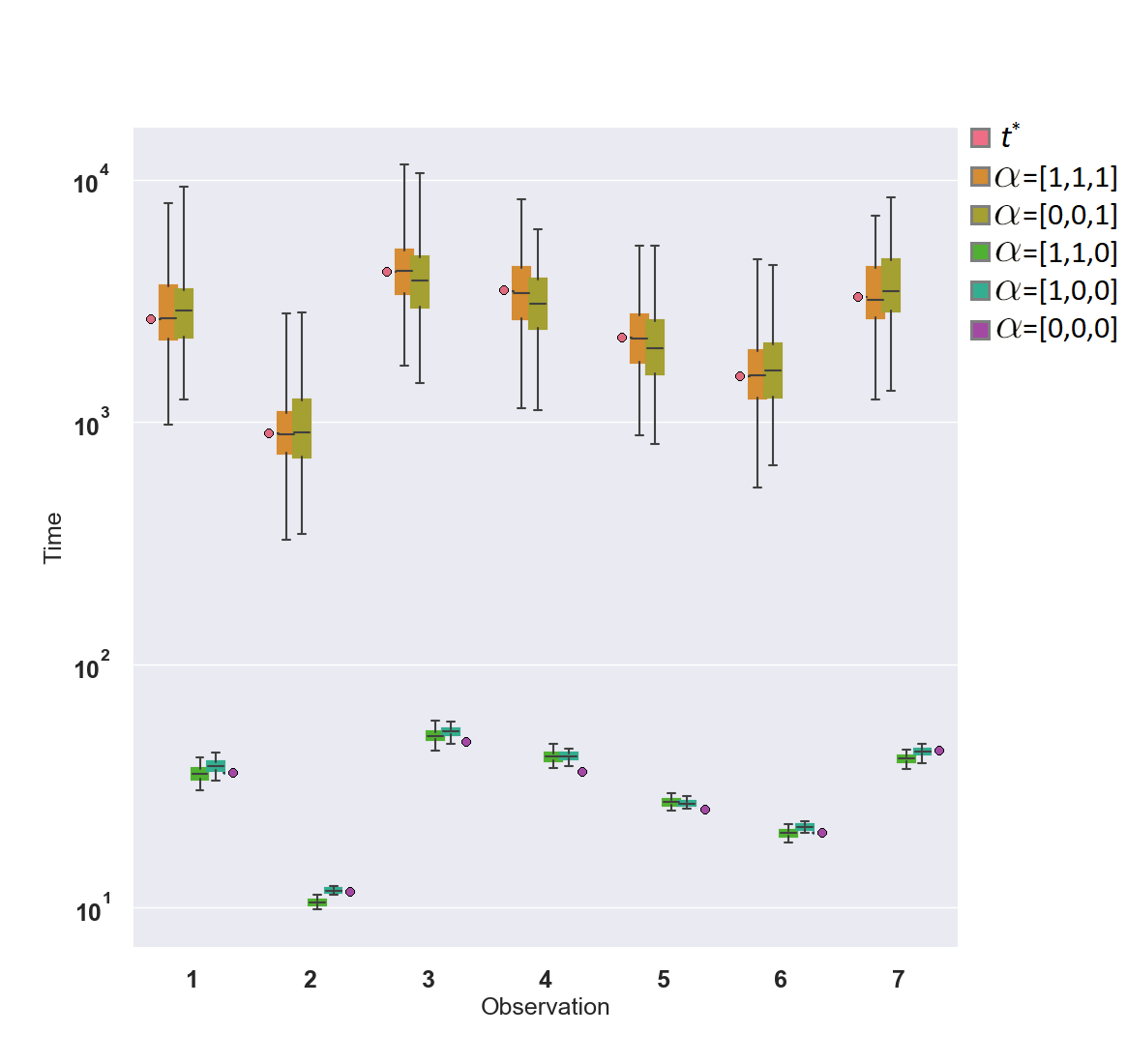}  
		& 
		\hspace{-3mm}
		\includegraphics[width=4.08cm]{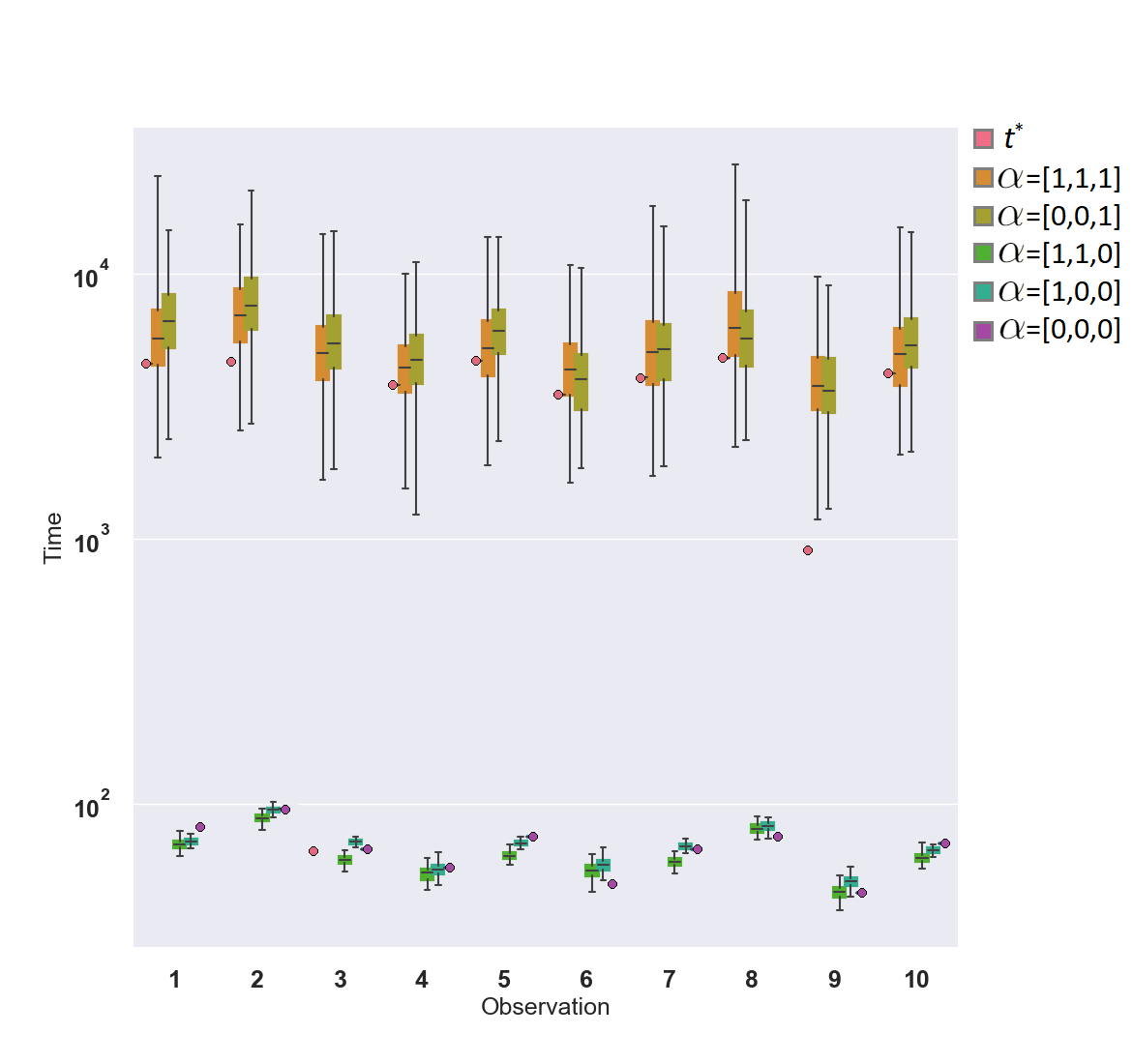}  &
		\hspace{-4mm}
		\includegraphics[height=3.65cm, width=4.08cm]{figures/support_U_0_1_obs__box_plot_noise}  
		& 
		\hspace{-3mm}
		\includegraphics[ width=4.08cm]{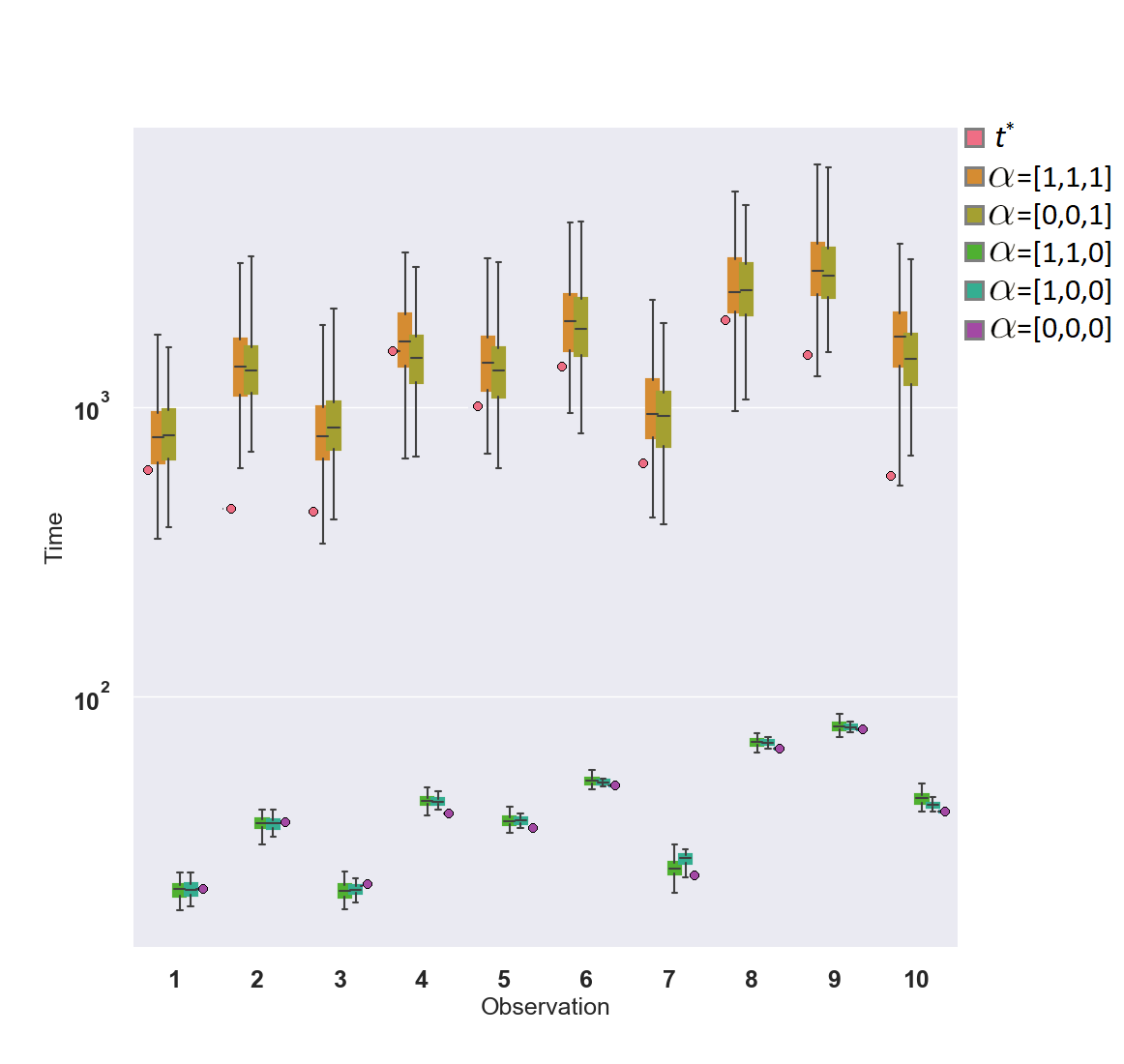}  \\
		(a) Non-Censored \hspace{-0mm} & \hspace{-5mm}
		(b) Censored &
		(c) Non-Censored \hspace{-0mm} & \hspace{-5mm}
		(d) Censored 
	\end{tabular}
	\vspace{-3mm}
	\caption{\small Effects of stochastic layers on uncertainty estimation on 10 randomly selected test-set subjects from the {\sc flchain} ( (a) and (b) ) and  {\sc support}  ( (c) and (d)) datasets. Ground truth times are denoted as $t^*$ and box plots represent time-to-event distributions from a 2-layer model, where $\bs{\alpha}=[\alpha_0,\alpha_1,\alpha_2]$ indicates whether the corresponding noise source, $\{\epsilonv_0,\epsilonv_1, \epsilonv_2\}$, is active. For example $\bs{\alpha}=[1,0,0]$ indicates noise on the input layer only.}
	\vspace{-5mm}
	\label{fig:noise}
\end{figure*}

Figure \ref{fig:noise} shows the contribution effects of stochastic layers for noise $\rm{Uniform}$(0,1)  on both censored and non-censored time-to-event distributions.  Tables \ref{tb:noise_rae} and \ref{tb:noise_ci} compares noise sources on relative absolute error and CI.

\begin{table}[h!]
	\begin{center}
		\begin{small}
			\caption{Introduced proportion of missing values comparison on Flchain relative absolute error. Ranges in parentheses are 50\% empirical ranges over (median) test-set predictions.}
			\label{tb:missing_rae}
			\resizebox{0.5\textwidth}{!}{
				\begin{tabular}{ lrrrr}
					\toprule
					&0.10 &  0.20 & 0.30 & 0.50 \\
					\hline
					Non-Censored &&&\\
					\hline
					DATE &19.9$_{(9.6 ,32.7)}$&  {\bf19.8}$_{(9.1,33.7)}$&{\bf19.7}$_{(10.8,33.2)}$& 19.7$_{( 10.3,  33.5)}$\\
					DATE-AE  & {\bf19.2}$_{(9.6 ,34.9)}$&21.9$_{((9.5,33.4)}$&20.6$_{(9.7,32.8)}$& {\bf18.3}$_{( 9.5,  32.9)}$\\
					DRAFT  &32.9$_{(10.0,92.3)}$& 34.1$_{(11.5 ,119.8)}$&{\bf19.7}$_{(10.3, 33.5)}$&19.7$_{( 10.3,33.5)}$\\
					\hline
					Censored &&&\\
					\hline
					DATE  &{\bf0}$_{(0,20.4)}$& 1.9$_{(0,19.4)}$& 2.7$_{( 0,20.1)}$&7.3$_{(0,21.8)}$\\
					DATE-AE &  {\bf0}$_{( 0, 12.9)}$&  3$_{( 0,19)}$&{\bf2.1}$_{(0, 16.5)}$& {\bf6}$_{( 0,21.3)}$\\
					DRAFT  & {\bf0}$_{( 0, 0)}$& {\bf0}$_{(0,0)}$&7.3$_{(0, 21.8)}$& 7.3$_{(0, 21.8)}$\\
					\bottomrule
				\end{tabular}
			}
		\end{small}
	\end{center}
\vspace{-5mm}
\end{table}	
\begin{table}[h!]
	\begin{center}
		\begin{small}
			\caption{ Introduced proportion of missing values comparison on  {\sc flchain} Concordance-Index.}
			\label{tb:missing_ci}
			\begin{tabular}{ lrrrr}
				\toprule
				&0.10 &  0.20 & 0.30 & 0.50 \\
				\midrule 
				DATE  &0.815&0.803& {\bf 0.803} &0.784\\
				DATE-AE  &0.814&0.804&0.799& {\bf 0.785}\\
				DRAFT  &{\bf0.822}& {\bf 0.807}&0.801& 0.783\\
				\bottomrule
			\end{tabular}
		\end{small}
	\end{center}
\end{table}	

\begin{table}[h!]
	\begin{center}
		\begin{small}
			\caption{Concordance-Index results on test data.}
			\label{tb:c-index}
			\resizebox{0.99\columnwidth}{!}{
			\begin{tabular}{lrrrrr}
				& DATE & DATE-AE & DRAFT &  Cox-Efron & RSF \\
				\hline
				{\sc ehr} &{\bf0.78}  &{\bf0.78} &0.76 &0.75&--\\
				{\sc flchain} &  {\bf0.83} & {\bf0.83} &  {\bf0.83} & {\bf0.83} & 0.82 \\
				{\sc support}  & 0.84&0.83&{\bf 0.86}&0.84& 0.80 \\
				{\sc seer}  & {\bf0.83}&{\bf0.83}&{\bf0.83}&0.82& 0.82\\
				\hline
			\end{tabular}
		}
		\end{small}
	\end{center}
\vspace{-2mm}
\end{table}	

\begin{table}[h!]
	\centering
	\caption{Median relative absolute errors (as percentages of $t_{\rm max}$), on non-censored and censored data. Ranges in parentheses are 50\% empirical ranges over (median) test-set predictions.}
	\label{tb:rae}
	\resizebox{\columnwidth}{!}{
		\begin{tabular}{lrrr}
			& DATE & DATE-AE & DRAFT \\
			\hline
			Non-censored & & & \\
			\hline
			{\sc ehr} & {\bf 23.6}$_{(11.1,43.0)}$ & 24.5$_{(12.4,44.0)}$ & 36.7$_{(16.1,81.3)}$ \\
			{\sc flchain} & 19.5$_{(9.5,31.1)}$ & {\bf 19.3}$_{(8.9,32.4)}$ & 26.2$_{(9.0,53.5)}$ \\
			{\sc support} & 2.7$_{(0.4,16.1)}$ & {\bf 1.5}$_{(0.4,19.2)}$ & 2.0$_{(0.2,35.3)}$ \\
			{\sc seer} & {\bf 18.6}$_{(8.3,34.1)}$ & 20.2$_{(10.3,35.8)}$ & 23.7$_{(9.9,51.2)}$\\
			\hline
			Censored & & & \\
			\hline
			{\sc ehr} & 12.4 $_{(0,38.7 )}$&  1.6$_{(0,34.)}$ &  {\bf 0} $_{( 0, 0)}$ \\
			{\sc flchain} & 0$_{( 0 ,18.8)} $ &0$_{ (0 ,15.6)} $&  {\bf 0} $_{( 0, 0)}$ \\
			{\sc support} &0$_{( 0,13.0)}$  & 0$_{( 0,  8.8)} $&  {\bf 0} $_{( 0, 0)}$\\
			{\sc seer} & {\bf 0} $_{(0, 0)}$ & {\bf 0} $_{( 0, 0)}$&  {\bf 0} $_{( 0, 0)}$\\
			\hline
		\end{tabular}
	}
	\vspace{-4mm}
\end{table}

\begin{table}[h]
	\begin{center}
		\begin{small}
			\caption{\small Effects of noise source and stochastic layers on {\sc support} Median relative absolute error. Ranges in parentheses are 50\% empirical ranges over (median) test-set predictions. }
			\label{tb:noise_rae}
			\begin{tabular}{ lrrr}
				& {\rm Uniform}(-1,1) &  {\rm Uniform}(0,1) & {\rm Gaussian}(0,1 ) \\
				\hline
				Non-censored & & & \\
				\hline
				All &  2.4$_{(0.4,19.9)}$&  2.2$_{(0.5,  19.2)}$&   1.9$_{(0.4, 17.)}$\\
				Input &2.2$_{(0.4 ,18.)}$ & {\bf1.8}$_{(0.4,16.1)}$&   1.9$_{(0.4 , 14.9)}$\\
				Output && 2.6$_{(0.4 ,21.1)}$&\\
				\hline
				Censored & & & \\
				\hline
				All  &  {\bf0}$_{(0,14.6)}$& {\bf0} $_{(0,13.7)}$&  {\bf0} $_{( 0,16.4)}$\\
				Input  &{\bf0}$_{( 0,15.3)}$ & 1.2$_{(0, 22.4)}$& 0.8 $_{( 0,21.2)}$\\
				Output && {\bf0}$_{(0,8.2)}$&\\
				\hline
			\end{tabular}
		\end{small}
	\end{center}
\end{table}	
\begin{table}[h!]
	\begin{center}
		\begin{small}
			\caption{ \small Effects of noise source and stochastic layers on {\sc support} concordance-index.} 
			\label{tb:noise_ci}
			\begin{tabular}{ cccc}
				\toprule
				& {\rm Uniform}(-1,1) &  {\rm Uniform}(0,1) & {\rm Gaussian}(0,1 ) \\
				\midrule 
				All & 0.825& {\bf0.835}& 0.826\\
				Input & 0.841& 0.829&  0.825 \\
				Output &&{\bf0.836}&\\
				\bottomrule
			\end{tabular}
		\end{small}
	\end{center}
\end{table}	
%

\newpage

%
%
\begin{figure*}[t!]
	\centering
	\includegraphics[ width=0.6\columnwidth]{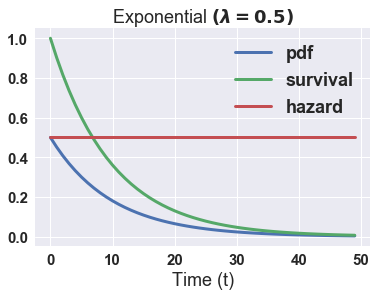}
	\includegraphics[ width=0.6\columnwidth]{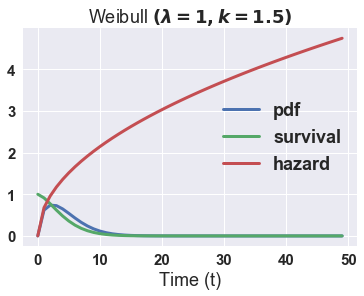}
	\includegraphics[ width=0.6\columnwidth]{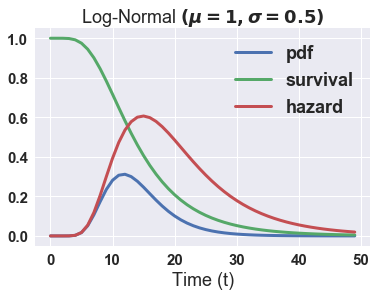}
	\vspace{-5mm}
	\caption{Popular parametric characterizations: exponential (left), Weibull (middle) and log-normal (right).}
	\label{fg:parametric_models}
\end{figure*}

%
\vspace{-0mm}
\section{Parametric examples of $f$, $h$ and $S$ relationships}
Figure \ref{fg:parametric_models} shows examples of exponential, Weibull and log-normal time-to-event pdf $f_T(t|\thetav)$ with corresponding survival function $S_T(t|\thetav)$ and $h(t| \thetav)$, where $\thetav$ are the pdf parameters  and $T$ is the time-to-event random variable. 


\vspace{-0mm}
\section{ Architecture of the neural network}
In all experiments, DATE and DRAFT are specified in terms of two-layer MLPs of $50$ hidden units with Rectified Linear Unit (ReLU) activation functions and batch normalization \cite{ioffe2015batch}.
The discriminator for DATE is a similarly defined MLP. 
As an optimizer, we use Adam \cite{kinga2015method} with the following hyperparameters: learning rate $3 \times 10 ^{-4}$, first moment $0.9$, second moment $0.99$, and epsilon $1 \times 10 ^{-8}$.
Further, we set the minibatch size to $M=350$ and use dropout with $p=0.8$ on all layers.
All the network weights are initialized using \textit{Xavier} \cite{glorot2010understanding}.
Datasets are split into training, validation and test sets as 80\%, 10\% and 10\% partitions, respectively, stratified by non-censored event proportion.
We use the validation set for early stopping and learning model hyperparameters.
DATE is executed using one NVIDIA P100 GPU with 16GB memory.

\bibliography{date}
\bibliographystyle{icml2018}